\documentclass[sigconf,nonacm,review=false]{acmart}

\usepackage{booktabs}
\usepackage{multirow}
\usepackage{colortbl}
\usepackage{subfigure}
\usepackage{mathrsfs}
\setcopyright{none}

\begin{document}

\title{Beyond Trajectory Statistics: Community-Consistent Human Mobility Generation via Evolving User–Location Graphs}


\author{Bo Liu}
\email{liubo317@hnu.edu.cn}
\affiliation{
  \department{National Supercomputing Center in Changsha}
  \institution{Hunan University}
  \city{Changsha}
  \country{China}
}

\author{Tong Li}
\email{tliay@hnu.edu.cn}
\affiliation{
  \department{National Supercomputing Center in Changsha}
  \institution{Hunan University}
  \city{Changsha}
  \country{China}
}

\author{Jiapeng Zhang}
\email{zhangjp@hnu.edu.cn}
\affiliation{
  \department{National Supercomputing Center in Changsha}
  \institution{Hunan University}
  \city{Changsha}
  \country{China}
}

\author{Zhu Xiao}
\email{zhxiao@hnu.edu.cn}
\affiliation{
  \department{National Supercomputing Center in Changsha}
  \institution{Hunan University}
  \city{Changsha}
  \country{China}
}

\author{Zhuo Tang}
\email{ztang@hnu.edu.cn}
\affiliation{
  \department{National Supercomputing Center in Changsha}
  \institution{Hunan University}
  \city{Changsha}
  \country{China}
}

\author{Kenli Li}
\email{lkl@hnu.edu.cn}
\affiliation{
  \department{National Supercomputing Center in Changsha}
  \institution{Hunan University}
  \city{Changsha}
  \country{China}
}
\begin{abstract}
Human mobility data are fundamental to Web-based location services and broader urban applications, and mobility generation provides synthetic trajectories when real data are costly or inaccessible. Although recent mobility generators have incorporated collective mobility patterns, they primarily focus on individual trajectories and aggregate population dynamics, leaving evolving community structures formed by user-location interactions insufficiently modeled. In this paper, we study community-consistent human mobility generation. To this end,
we formulate mobility generation as the generation of dynamic user-location graphs to model community structures and propose MOTION, a \textbf{Multi-agent framework} for cOmmuniTy-consIstent human mObility geNeration.
MOTION consists of \textbf{user agents}, \textbf{location agents}, and a global \textbf{manager agent}. MOTION first detects coherent communities from historical mobility using the Leiden algorithm and extracts community-conditioned interaction priors to guide the generation process. Guided by these priors, the user agents plan mobility, search for candidate destinations, and infer user movements from the individual view. In parallel, selectively activated location agents infer likely user arrivals from the community view. The manager coordinates both views, diagnoses interactions causing deviations, and provides targeted reflection signals to revise user decisions, location attraction, and source-user selection.
Experiments on two real-world datasets show that MOTION outperforms baselines by at least 4.05\% across reported metrics for individual trajectories, population and flow distributions, and community and graph structures. Further analyses confirm its robustness, privacy preservation, and downstream utility. The code is available at \url{https://anonymous.4open.science/r/MOTION-04C4}.
\end{abstract}



\maketitle
\thispagestyle{plain}
\pagestyle{plain}


\section{Introduction}

Web-based location services, including location-based social networks, point-of-interest (POI) recommendation, and online route planning~\cite{zhang2023deep,li2023artificial,xu2023urban}, rely on human mobility data to understand how users interact with locations~\cite{xu2024mmpoi,bae2024negative,liu2020polestar}. These interactions support personalized location recommendations and, when aggregated, provide valuable information for location demand analysis and other location-aware Web applications~\cite{liang2021fine,liang2019urbanfm,feng2018deepmove}. However, collecting and sharing user trajectories is often constrained by data scarcity, collection costs, and privacy concerns~\cite{xu2023urban,bassolas2019hierarchical,long2024physical,wang2024cola}. Human mobility generation therefore provides realistic synthetic trajectories to support these services while reducing reliance on sharing sensitive individual mobility records~\cite{wang2024cola,zhang2025noise,pellungrini2017data}.

\begin{figure}[tb]
    \centering
    \includegraphics[width=\linewidth]{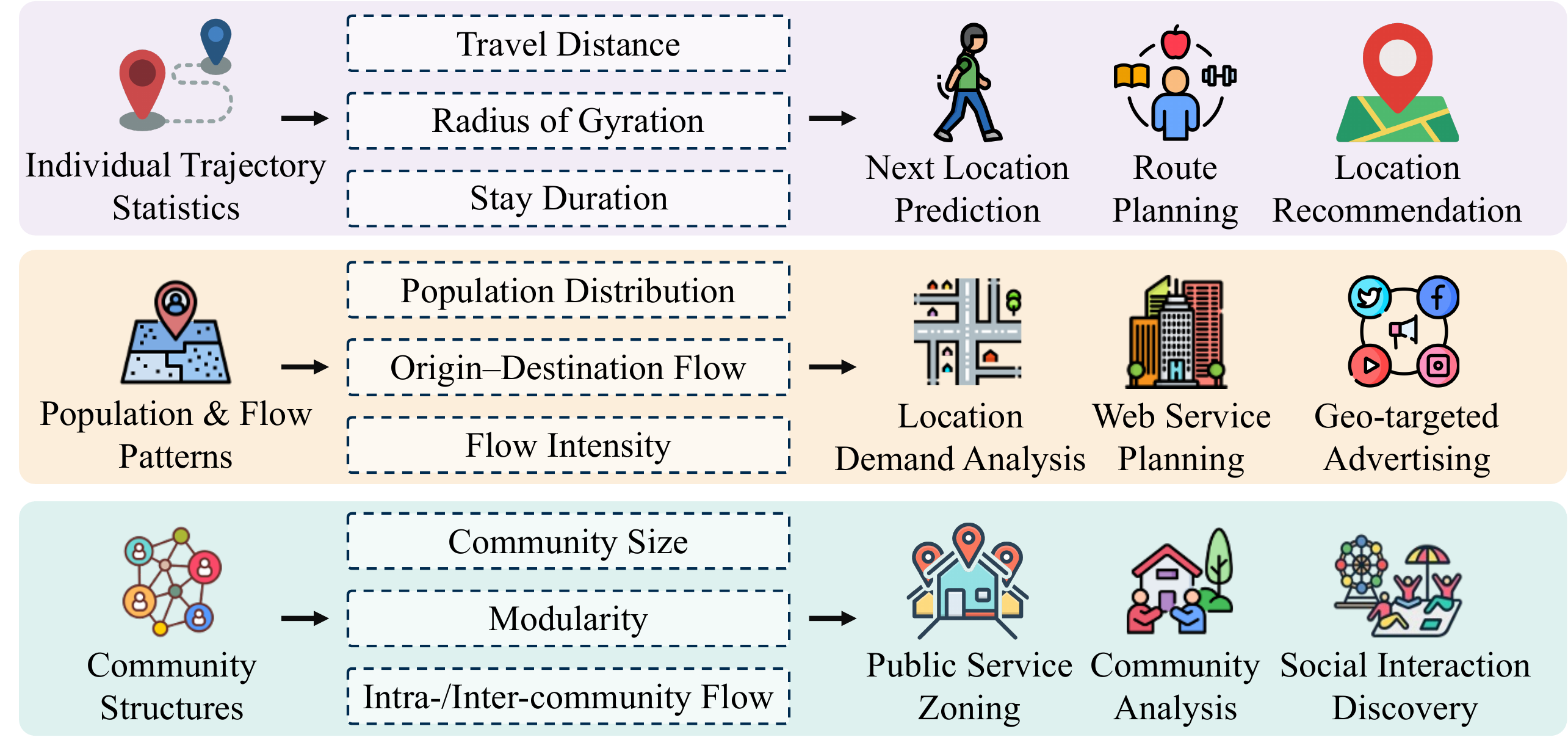}
    \vspace{-8mm}
    \caption{Three layers of human mobility patterns, their properties, and potential downstream applications.}
    \label{fig:intro}
\vspace{-6mm}
\end{figure}

To support Web applications, synthetic mobility should preserve three layers of mobility patterns~\cite{deng2025revisiting,rong2024interdisciplinary,you2018graphrnn,bojchevski2018netgan}. As illustrated in Fig.~\ref{fig:intro}, at the individual layer, trajectory statistics characterize user mobility preferences, supporting personalized POI recommendation and route planning~\cite{feng2020learning,luo2021stan}. At the population and flow layer, spatial distributions and origin-destination flows characterize aggregate mobility demand, supporting location demand analysis and geo-targeted Web services~\cite{liang2021fine,liang2019urbanfm}. At the community structure layer, shared destinations and recurring user-location interactions reveal mesoscopic community organization, providing a basis for community analytics, and social interaction discovery~\cite{ji2023community,li2020evolvegraph,yuan2012discovering}.

Existing studies have explored human mobility generation from both individual and collective perspectives. Mechanism-based methods~\cite{pappalardo2018data} and deep learning models~\cite{feng2020learning,chu2024simulating} capture individual routines, transition preferences, and spatiotemporal dependencies. Large language model (LLM)-based methods~\cite{lan2025open,yu2024harnessing,wang2024large,gong2024mobility,shao2024chain} further model activity semantics and destination choices. At the collective level, d-EPR~\cite{pappalardo2016human} exploits population-level location attractiveness, AgentMove~\cite{feng2025agentmove} extracts shared mobility patterns across users, MobAgent~\cite{li2024more} models heterogeneous population profiles, and CoDiffMob~\cite{zhang2025noise} and DeepMobility~\cite{yuan2025learning} incorporate population dynamics and collective flows into mobility generation.
Despite these advances, existing methods primarily emphasize individual trajectory realism and aggregate population and flow patterns, leaving the evolving community structures formed by user-location interactions insufficiently modeled. Such aggregate patterns do not fully capture which users repeatedly visit shared locations, how these interactions organize communities, or how community relationships evolve over time~\cite{li2020evolvegraph,ji2023community}. Therefore, mobility generation should go beyond matching trajectory statistics to explicitly preserve evolving community structures.

In this paper, we study community-consistent human mobility generation. To model community structures, we formulate mobility generation as dynamic user-location graph generation, where users and locations are represented as nodes and their interactions as edges. This unified representation explicitly captures the dependencies among users, locations, and mobility decisions, tracking user-location interactions and the resulting community structures over time~\cite{mauro2022generating,you2018graphrnn}. Achieving this objective requires addressing the following three challenges:

$\bullet$ \textbf{How to represent dynamic community structures for mobility generation?}
Community-consistent mobility generation requires priors that characterize how community structures and interaction patterns evolve with user-location interactions. These priors should preserve temporal and structural regularities without reproducing exact historical realizations. The first challenge is therefore to identify temporally coherent communities and extract representative interaction priors for generation.

$\bullet$ \textbf{How to generate user-location interactions efficiently for large-scale mobility datasets?}
Under the graph-based formulation, each mobility decision creates an interaction between a user and a location, which should reflect both the user's mobility intention and the location's role. However, reasoning over all possible user-location pairs is computationally prohibitive, potentially requiring over one million inferences to generate a single day of hourly mobility for only 100 users. The second challenge is therefore to efficiently consider complementary user-side and location-side evidence.

$\bullet$ \textbf{How to align generated user-location graphs with realistic mobility structures?}
Even when individual user-location interactions are plausible, their aggregation may still deviate from realistic community structures. Correcting such deviations requires identifying graph-level discrepancies, tracing them to the responsible user and location decisions, and determining how these decisions should be revised. The third challenge is therefore to translate structural deviations into targeted revisions of individual decisions.

To address these challenges, we propose MOTION, a Multi-agent framework for cOmmuniTy-consIstent human mObility geNeration. MOTION consists of \textbf{user agents} for target users, \textbf{location agents} for locations, and a global \textbf{manager agent}. Rather than learning previously unknown collective patterns through a trained predictor, MOTION extracts community-conditioned priors from historical mobility to guide generation and targeted reflection.
Specifically, MOTION first constructs dynamic user-location graphs from historical trajectories and applies Leiden-based dynamic community detection to identify evolving communities. Within each community, MOTION employs metagraph-based temporal template matching to extract recurring mobility motifs, population and flow dynamics, and co-mobility patterns. These priors capture temporal and structural regularities without reproducing exact historical realizations, serving as guidance for generation and structural targets for reflection, addressing the \textbf{first} challenge.
Guided by these priors, user agents and location agents generate user-location interactions from the individual and community views, respectively. Each user agent constructs a mobility plan based on mobility motifs, searches for candidate destinations, and infers movements using narrowed candidate-location contexts and reflection signals. In parallel, an active location selection mechanism activates relevant location agents based on population and flow deviations and moving users in nearby locations. Each activated location agent infers likely arriving users based on surrounding user contexts, population and flow dynamics, co-mobility patterns, and reflection signals. The two types of agents construct complementary individual-view and community-view graphs, addressing the \textbf{second} challenge.
The manager agent coordinates these proposals based on their supporting evidence to construct an intermediate graph. It then identifies deviations from realistic community structures and population and flow patterns, traces them to responsible user and location decisions, and generates targeted reflection signals to revise user mobility plans, destination preferences, location attraction estimates, and source-user selections. Only affected agents regenerate their proposals, addressing the \textbf{third} challenge.

The main contributions can be summarized as follows:

$\bullet$ We study community-consistent human mobility generation and formulate it as dynamic user-location graph generation to explicitly model evolving community structures.

$\bullet$ We propose MOTION, a prior-guided multi-agent framework that coordinates user and location agents to generate individual- and community-view graphs, and employs manager-driven targeted reflection to preserve structural consistency.

$\bullet$ We conduct extensive experiments on two real-world datasets, demonstrating that MOTION outperforms baselines by at least 4.05\% across the reported metrics. Additional analyses evaluate its robustness, privacy preservation, and downstream utility.

\begin{figure}[tb]
    \centering
    \subfigure[Real Data]{
        \includegraphics[width=0.30\columnwidth]{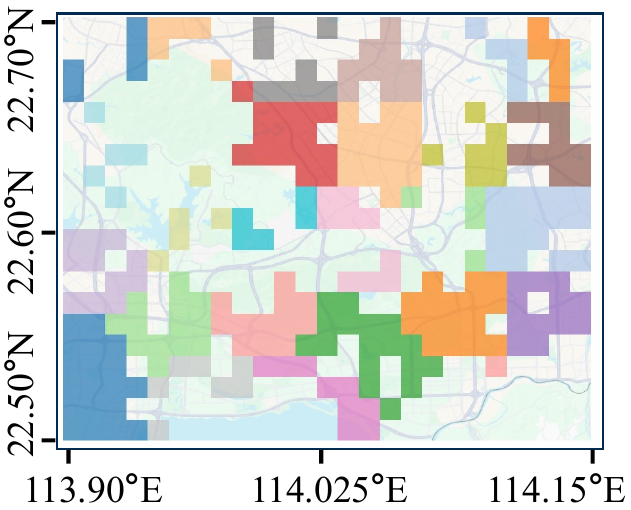}
    }\hfill
    \subfigure[CoDiffMob]{
        \includegraphics[width=0.30\columnwidth]{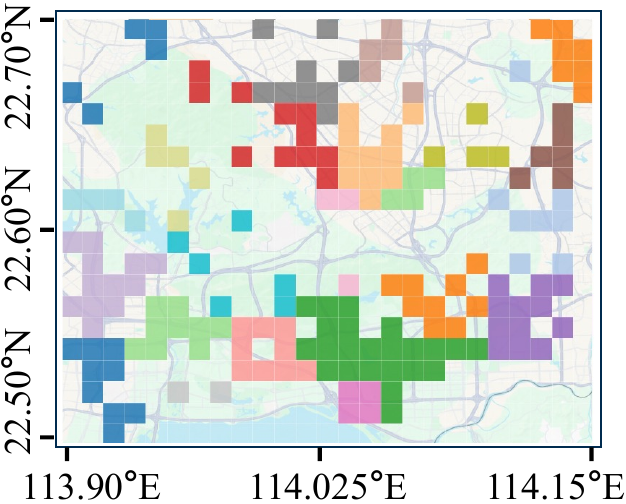}
    }\hfill
    \subfigure[LLMob]{
        \includegraphics[width=0.30\columnwidth]{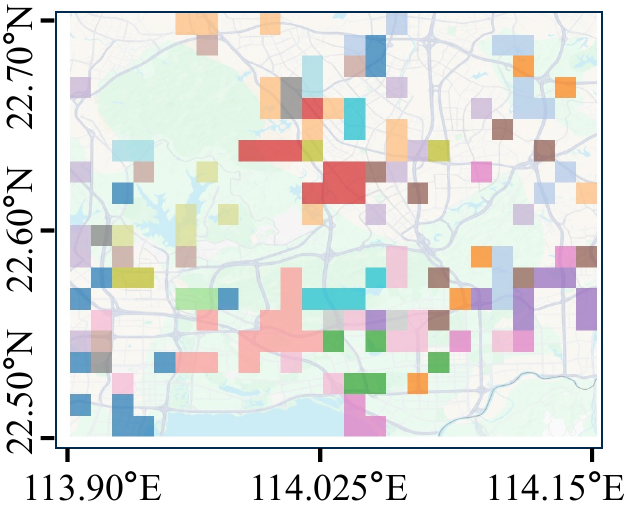}
    }
    \vspace{-5mm}
    \caption{Community structures detected from real and generated mobility. Different colors represent different communities on the same gridded map.}
    \label{fig:preliminary_community}
    \vspace{-4mm}
\end{figure}

\begin{figure*}[t]
    \centering
    \includegraphics[width=0.9\textwidth]{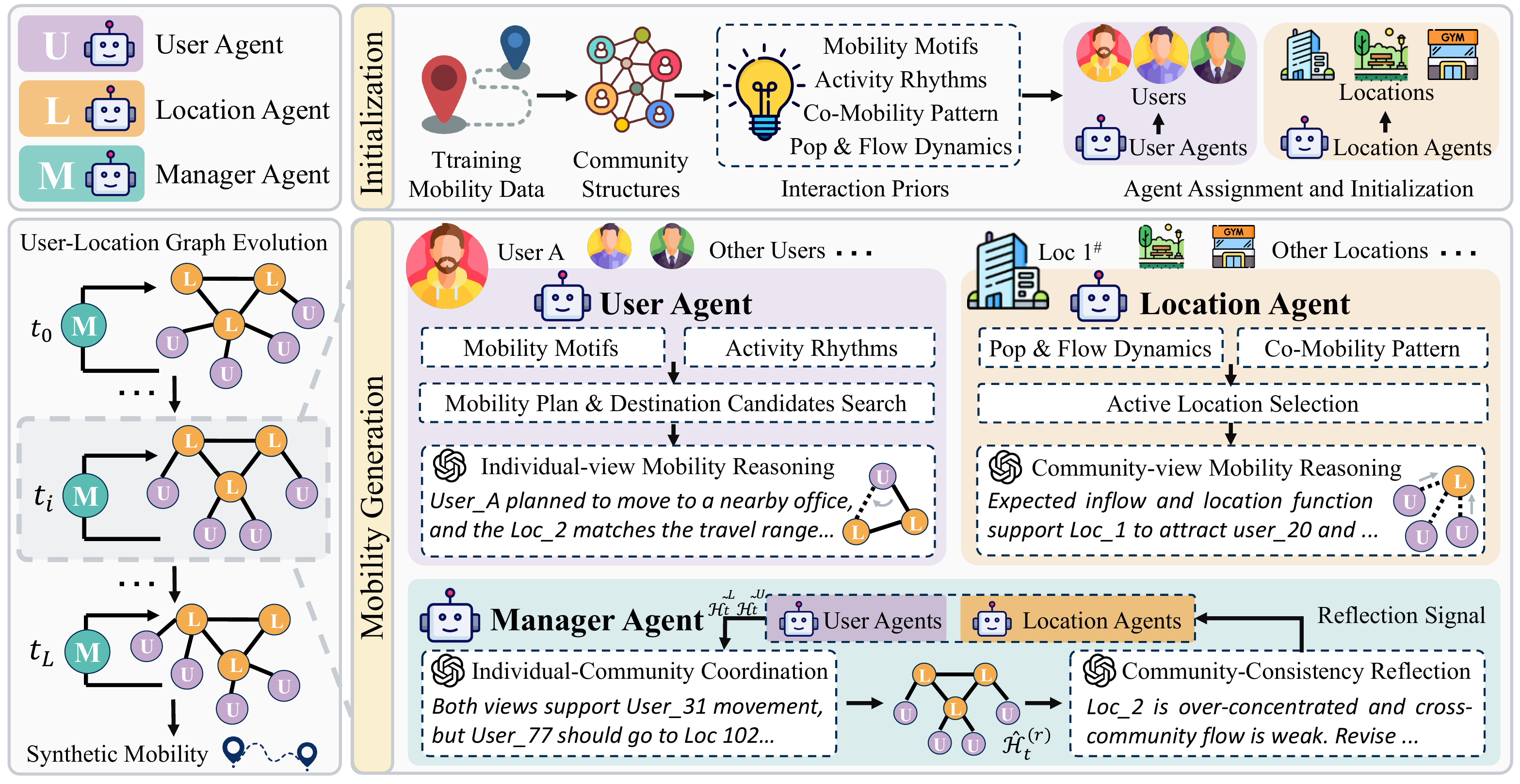}
    \vspace{-3mm}
    \caption{Overview of MOTION\@. MOTION extracts community structures and interaction priors from historical mobility to guide user- and location-agent generation from individual and community views. The manager coordinates their proposals and performs prior-guided targeted reflection to improve structural consistency.}
    \vspace{-3mm}
    \label{fig:framework}
\end{figure*}

\section{Preliminary Study}
\subsection{Community Structural Consistency}
A mobility community consists of users and locations connected through synchronized movements and interdependent population flows~\cite{li2020evolvegraph}, capturing mesoscopic activity regions and interaction boundaries~\cite{ji2023community}. Although recent mobility generators incorporate collective mobility patterns, preserving community structures formed by user-location interactions remains challenging.
To illustrate this limitation, we compare real mobility with CoDiffMob~\cite{zhang2025noise} and LLMob~\cite{wang2024large}, which model collective mobility and use LLMs, respectively. We apply the same dynamic community detection procedure to all datasets and visualize the detected communities on the same gridded urban map. As shown in Fig.~\ref{fig:preliminary_community}, real mobility exhibits large, spatially coherent communities. CoDiffMob recovers some coherent regions but produces more fragmented community assignments, whereas LLMob yields more scattered communities.
These observations suggest that modeling individual and aggregate mobility patterns does not necessarily preserve community-level organization. Therefore, mobility generation should explicitly model and preserve evolving community structures.

\subsection{Problem Definition}
Given real-world mobility data and the initial mobility states of a set of target users, human mobility generation aims to generate realistic future mobility trajectories.
Mathematically, let $\mathcal{U}=\{u_i\}_{i=1}^{N}$ and $\mathcal{L}=\{\ell_j\}_{j=1}^{M}$ denote the user and location sets. The historical mobility dataset is $\mathcal{S}=\{s_i\}_{i=1}^{N}$, where $s_i=[(\ell_0^i,t_0^i),\ldots,(\ell_{T_i}^i,t_{T_i}^i)]$ is the trajectory of user $u_i$. Given a set of target users $\hat{\mathcal{U}}=\{\hat{u}_i\}_{i=1}^{N^*}$ and their initial states $\mathcal{X}_0=\{(\ell_0^i,t_0^i)\}_{i=1}^{N^*}$, let $\mathcal{G}$ denote a mobility generation model. The generated trajectories are formulated as:
\begin{equation}
\hat{\mathcal{S}} = \{\hat{s}_i\}_{i=1}^{N^*} = \mathcal{G}(\mathcal{S},\mathcal{X}_0).
\end{equation}
 
\section{Methodology}
\subsection{Framework Overview}
The overall framework of MOTION is illustrated in Fig.~\ref{fig:framework}. MOTION consists of a \textbf{user agent} for each target user, a \textbf{location agent} for each location, and a global \textbf{manager agent}. During initialization, MOTION constructs dynamic user-location graphs from historical trajectories, detects coherent communities using temporally smoothed Leiden-based community detection, and extracts community-conditioned interaction priors using metagraph-based mining. Rather than training a predictor to discover new collective patterns, MOTION uses these priors to guide agent decisions and provide structural reference targets for reflection.
During generation, user agents construct mobility plans, search for candidate destinations, and infer movements from the individual view. Meanwhile, an active location selection mechanism activates structurally relevant location agents to infer likely user arrivals from the community view. The manager coordinates these complementary proposals, identifies structural deviations from the extracted priors, and traces them to responsible user and location decisions. It then provides targeted reflection signals for the affected agents to revise their proposals, and the refined graph conditions the next time step.

\subsection{Mobility Prior Initialization}
To represent the interaction and community structures emerging across users, MOTION models historical mobility as dynamic user-location graphs, and then extracts temporally coherent communities and community-conditioned interaction priors.

\subsubsection{Dynamic User-Location Graph Construction}
MOTION represents human mobility as a sequence of dynamic user-location graph snapshots. Each snapshot contains time-varying user-location edges indicating users' current locations and static location-location edges capturing geographic proximity and historical mobility connectivity.
Specifically, let $\mathcal{U}$ and $\mathcal{L}$ denote the historical user and location sets. After discretizing trajectories into $T$ time intervals, MOTION constructs:
\begin{equation}
    \mathbb{H}=\{\mathcal{H}_1,\ldots,\mathcal{H}_T\}, \qquad
    \mathcal{H}_t=
    \left(
    \mathcal{U}\cup\mathcal{L},
    \mathcal{E}^{UL}_t\cup\mathcal{E}^{LL}
    \right),
\end{equation}
where $(u_i,\ell_j)\in\mathcal{E}^{UL}_t$ indicates that user $u_i$ is located at $\ell_j$ at time $t$. The static edge set $\mathcal{E}^{LL}$ connects locations according to geographic proximity and aggregate historical flows, providing a spatial backbone shared across snapshots. Changes from $\mathcal{E}^{UL}_{t-1}$ to $\mathcal{E}^{UL}_t$ encode individual movements and jointly determine location populations and inter-location flows. Therefore, human mobility is represented as sequential user-location graphs rather than isolated location transitions. 

\subsubsection{Community-Conditioned Interaction Prior Mining}
To capture community structures and interaction patterns, MOTION employs temporally smoothed Leiden-based community detection. Let $\mathcal{C}_t$ denote the community partition of $\mathcal{H}_t$. For $t>1$, Leiden optimizes the Constant Potts Model (CPM) objective~\cite{traag2019louvain}, initialized with the preceding partition $\mathcal{C}_{t-1}^{*}$. Temporal continuity is further encouraged by stochastically retaining each location's previous community assignment, while allowing memberships to evolve with user-location interactions. Each user is then assigned to the community of its current location.
Based on the detected communities, MOTION employs metagraph-based temporal template matching to extract community-conditioned interaction priors. Individual-level priors include mobility motifs, motif transitions, and activity rhythms, while community-level priors capture community memberships and sizes, population and flow dynamics, and co-mobility patterns. Together, these priors characterize how individual movements shape evolving community structures and guide subsequent generation. Representative examples are provided in Appendix~\ref{appendix:prior_mining}.

\subsection{User Agent}
MOTION assigns a user agent to each target user. Each agent maintains the user's mobility state, constructs a daily mobility plan, searches for candidate destinations, and infers the next location from the individual view.

\subsubsection{Individual Mobility Planning}
To organize daily mobility decisions, the user agent first samples a community for user $\hat{u}_i$ from the affiliation distribution $p(c\mid\ell_0^i)$ extracted from historical mobility. The user agent then samples a sequence of mobility motifs conditioned on the assigned community, time of day, and motif transition patterns. The transition between consecutive motifs is modeled as:
\begin{equation}
    m_{r+1}^i \sim
    p\left(m_{r+1}\mid m_r^i,c_i,h_r\right),
\end{equation}
where $m_r^i$ denotes the current motif, $c_i$ denotes the assigned community, and $h_r$ denotes its activation time.
Each motif is instantiated as a sequence of stay and move actions based on extracted activity rhythms, destination semantics, travel-distance ranges, and return patterns. The resulting plan specifies movement timing, destination intentions, and the expected movement structure. Users with compatible plans and community memberships are further associated through co-mobility patterns to support location-side reasoning. During generation, targeted reflection may adjust planned actions, execution times, travel ranges, or destination intentions.

\subsubsection{Individual-View Mobility Reasoning}
At each time step, the user agent executes the active segment of its mobility plan to determine the next location. For a stay action, the current user-location edge is retained. For a return action, the agent revisits a recorded location to preserve the mobility motif. To reason efficiently and reliably about other movements, MOTION searches for candidate destinations within the planned travel-distance range and evaluates their semantic compatibility, geographic distance, regional attraction, and community flow evidence. The user agent jointly interprets these signals together with reflection signals to select the next location. The decisions of all user agents are aggregated into the individual-view graph $\widetilde{\mathcal{H}}_t^{U}$.

\begin{figure}[t]
    \centering
    \vspace{-3mm}
    \includegraphics[width=\columnwidth]{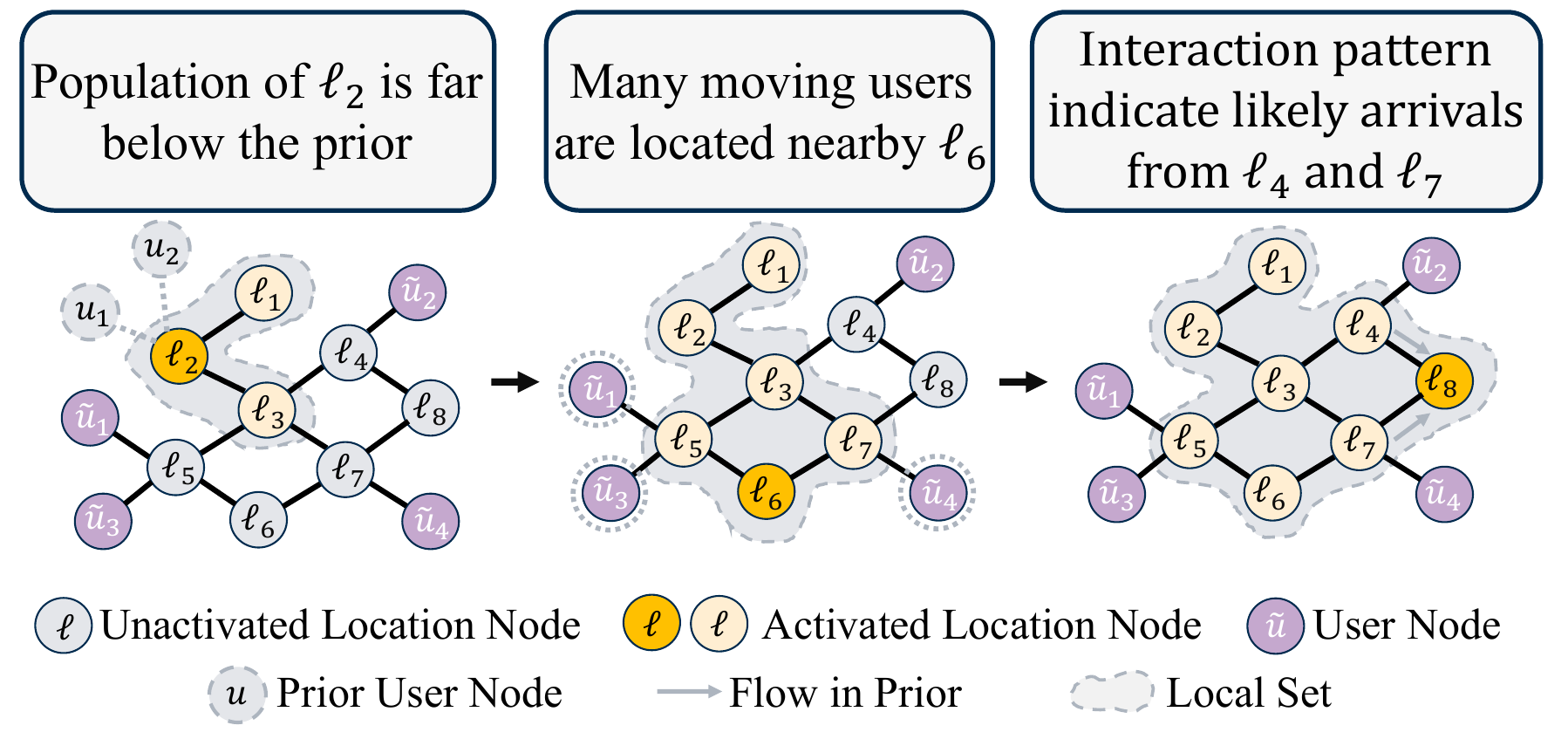}
    \vspace{-8mm}
    \caption{Local-set formation. Locations and their neighborhoods are activated according to population dynamics, nearby moving users, and prior-supported incoming flows and are then merged into a local set.}
    \vspace{-3mm}
    \label{fig:location_selection}
\end{figure}

\subsection{Location Agent}
MOTION assigns a location agent to each location to model mobility from the receiver-side community view. An active location selection mechanism selectively activates relevant location agents to infer likely user arrivals. This receiver-side reasoning complements user agents' destination decisions and promotes realistic collective interaction patterns.

\subsubsection{Active Location Selection}
To improve reasoning efficiency, MOTION identifies active locations based on deviations between current and expected populations, moving users in nearby or graph-connected source locations, and prior-supported incoming flows and co-mobility patterns. MOTION then groups these locations and their neighbors into local sets, as illustrated in Fig.~\ref{fig:location_selection}. From each local set, MOTION randomly samples at most \texttt{active\_m} locations for agent activation. The sampled locations are merged into the final active location set. This strategy ensures coverage across local mobility regions while preventing excessive activation of popular locations.

\subsubsection{Community-View Mobility Reasoning}
At each time step, each activated location agent identifies candidate users from graph-connected source locations and infers their potential arrivals. The location agent considers users' current locations, mobility plans, activity intentions, community memberships, and co-mobility relations, together with the target location's semantic role, expected population, and incoming flows.
It then selects users whose arrivals align with plausible graph connections, activity intentions, and community-conditioned interaction patterns. Co-mobility evidence further supports synchronized movements toward shared destinations. If multiple locations select the same user, MOTION retains the proposal best supported by population, flow, and co-mobility priors. All location-agent proposals are aggregated into the community-view graph $\widetilde{\mathcal{H}}_t^{L}$ for subsequent coordination with the individual-view graph by the manager agent.

\subsection{Manager Agent}
The manager agent coordinates individual- and community-view graphs and uses extracted interaction priors to diagnose structural deviations and guide targeted reflection. This prior-guided process aims to preserve individual mobility intentions while improving consistency with realistic community structures.

\subsubsection{Individual-Community Coordination}
At each time step, the manager agent coordinates the individual-view graph $\widetilde{\mathcal{H}}_t^{U}$ and the community-view graph $\widetilde{\mathcal{H}}_t^{L}$. For each user, MOTION considers the user agent's mobility plan, movement history, intended activity, proposed destination, and rationale, together with the location agent's attraction estimates, expected population and flows, community role, co-mobility evidence, and rationale. When both views agree, their proposal is retained. Otherwise, the manager uses LLM reasoning to evaluate the supporting evidence and resolve conflicts by balancing individual intentions with community-conditioned interaction priors. The resolved interactions form an intermediate graph $\widehat{\mathcal{H}}_t^{(r)}$ at coordination round $r$, which is subsequently evaluated for structural consistency.

\begin{figure}[tb]
    \centering
    \vspace{-3mm}
    \includegraphics[width=\columnwidth]{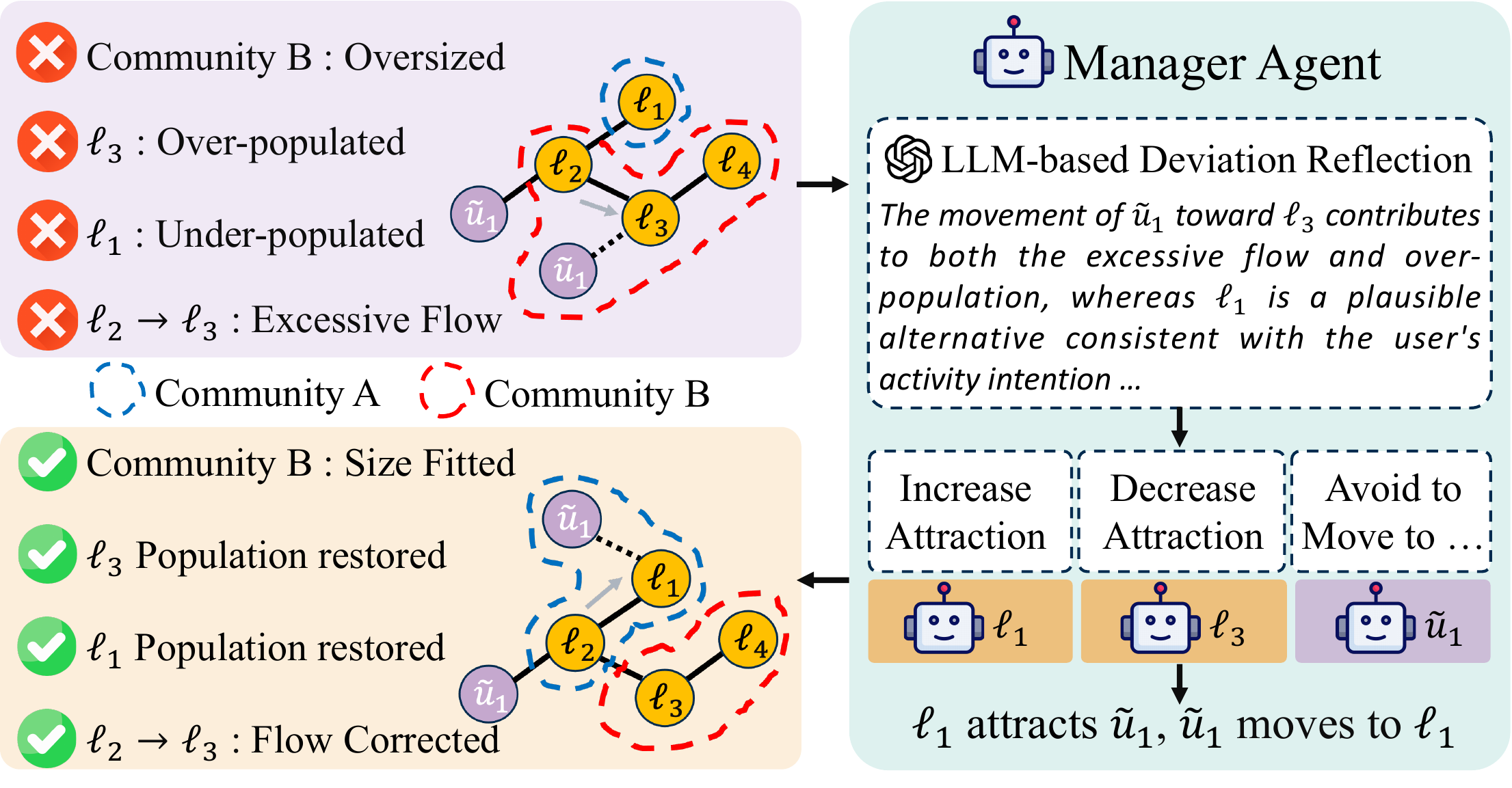}
    \vspace{-8mm}
    \caption{Community-consistency reflection. MOTION bridges generated and real communities, diagnoses deviations, attributes them to related decisions, and sends reflection signals to regenerate the affected proposals.}
    \vspace{-5mm}
    \label{fig:manager_reflection}
\end{figure}

\subsubsection{Community-Consistency Reflection}
Even individually plausible interactions may collectively deviate from realistic population, flow, and community structures. MOTION therefore iteratively coordinates agent proposals and refines the generated graph using historical interaction priors:
\begin{equation}
\begin{aligned}
\widehat{\mathcal{H}}_t^{(r)}
&=
\operatorname{Coord}\left(
\widetilde{\mathcal{H}}_t^{U,(r)},
\widetilde{\mathcal{H}}_t^{L,(r)}
\right),\\
\widehat{\mathcal{H}}_t^{(r+1)}
&=
\operatorname{Reflect}\left(
\widehat{\mathcal{H}}_t^{(r)},
\mathcal{P}_t
\right),
\end{aligned}
\end{equation}
where $\mathcal{P}_t$ denotes community-conditioned priors extracted from historical mobility, and $\operatorname{Reflect}$ represents prior-guided diagnosis, targeted agent regeneration, and re-coordination.
To identify structural deviations, MOTION applies the same temporally smoothed community detection procedure to the intermediate graph and aligns generated communities with time-corresponding historical reference communities using location-node Jaccard similarity. It then compares community memberships and sizes, location populations, and intra- and inter-community flows against the extracted priors.
As illustrated in Fig.~\ref{fig:manager_reflection}, MOTION identifies locations with mismatched community memberships or substantial population and flow deviations. For each deviating location, the manager examines users staying at, departing from, or arriving at that location, along with flow-connected source and destination locations, to identify the responsible interaction decisions.
Based on this diagnosis, the manager provides targeted reflection signals to the corresponding agents. User agents may revise planned actions, execution times, travel ranges, or destination intentions, while location agents may adjust attraction estimates, preferred source locations, or candidate users. Only the affected agents regenerate their proposals, which are then re-coordinated to obtain $\widehat{\mathcal{H}}_t^{(r+1)}$.

\begin{table}[tb]
\centering
\caption{Evaluation metrics for the three pattern layers.}
\vspace{-2mm}
\footnotesize
\label{tab:evaluation_metrics}
\setlength{\tabcolsep}{4pt}
\renewcommand{\arraystretch}{1.08}
\begin{tabular}{>{\centering\arraybackslash}m{\dimexpr0.5\columnwidth-2\tabcolsep-0.5\arrayrulewidth\relax}|>{\centering\arraybackslash}m{\dimexpr0.5\columnwidth-2\tabcolsep-0.5\arrayrulewidth\relax}}
\toprule
\rowcolor[gray]{0.92}
\textbf{Pattern Layer} & \textbf{Evaluation Metrics} \\
\midrule
Individual Trajectory Statistics & Distance, Radius, LocNum \\
\midrule
Population and Flow Distributions & Pop-EMD, Flow-EMD, G-Rank \\
\midrule
\begin{tabular}[c]{@{}c@{}}
Community and Graph-\\
Structural Consistency
\end{tabular}
&
\begin{tabular}[c]{@{}c@{}}
Motif, Comm-Size,\\
Comm-Mem, Comm-Trans
\end{tabular}
\\
\bottomrule
\end{tabular}
\end{table}

\section{Experiments}
\subsection{Experimental Setup}
\subsubsection{Datasets and Experimental Settings}
We evaluate MOTION on two real-world datasets. The \textbf{private car dataset}~\cite{liu2026multi} contains two weeks of GPS traces from 2,342 private vehicles, while the \textbf{mobile phone dataset}~\cite{shao2024chain} contains two weeks of GPS traces from 5,352 users. Both datasets were collected with explicit authorization from users and data providers and were de-identified.
We adopt an 8:2 user-level train-test split and reserve the final day for evaluation. Train and test users are disjoint, and test users' historical trajectories are excluded from prior extraction or training. We employ GPT-4o as the default LLM backbone for MOTION and baselines. Dataset statistics, ethical information, and POI profiles are provided in the supplementary material; detailed settings appear in Appendix~\ref{appendix:settings}.

\subsubsection{Baselines and Metrics}
We compare MOTION with seven representative mobility generators, including the GAN-based model MoveSim~\cite{feng2020learning}, the collaborative generator DeepMobility~\cite{yuan2025learning}, the diffusion-based model CoDiffMob~\cite{zhang2025noise}, LLM-based methods LLMob~\cite{wang2024large} and LLM-COD~\cite{yu2024harnessing}, and the agent-based models TrajLLM~\cite{ju2025trajllm} and AgentMove~\cite{feng2025agentmove}.
Detailed descriptions of the baselines are provided in Appendix~\ref{appendix:baselines}.
We evaluate MOTION at the three pattern layers summarized in Table~\ref{tab:evaluation_metrics}. Detailed metric definitions are provided in Appendix~\ref{appendix:metrics}.

\begin{table*}[t]
\centering
\caption{Overall performance on the private car and mobile phone datasets.}
\vspace{-3mm}
\label{tab:overall_performance}
\setlength{\tabcolsep}{2.3pt}
\scriptsize
\newcommand{\std}[1]{\ensuremath{_{\pm #1}}}
\resizebox{0.98\textwidth}{!}{%
\begin{tabular}{c|ccc|ccc|cccc}
\toprule
\multirow[c]{2}{*}{\raisebox{-0.3\height}{\textbf{Methods}}}
& \multicolumn{3}{c|}{\textbf{Individual Trajectory Statistics}}
& \multicolumn{3}{c|}{\textbf{Population and Flow Distributions}}
& \multicolumn{4}{c}{\textbf{Community and Graph-Structural Consistency}} \\
\cmidrule(lr){2-4}\cmidrule(lr){5-7}\cmidrule(lr){8-11}
& \textbf{Distance}$\downarrow$ & \textbf{Radius}$\downarrow$ & \textbf{LocNum}$\downarrow$
& \textbf{Pop-EMD}$\downarrow$ & \textbf{Flow-EMD}$\downarrow$ & \textbf{G-Rank}$\downarrow$
& \textbf{Motif}$\downarrow$ & \textbf{Comm-Size}$\downarrow$ & \textbf{Comm-Mem}$\uparrow$ & \textbf{Comm-Trans}$\uparrow$ \\
\midrule
\multicolumn{11}{c}{\textbf{Private Car Dataset}} \\
\midrule
MoveSim
& 0.2257\std{0.0034} & 0.1781\std{0.0029} & 0.2575\std{0.0038}
& 0.1037\std{0.0020} & 0.1014\std{0.0019} & 0.3475\std{0.0051}
& 0.2918\std{0.0044} & 0.2203\std{0.0035} & 0.2756\std{0.0033} & 0.2334\std{0.0007} \\
DeepMobility
& 0.2086\std{0.0053} & 0.1185\std{0.0019} & 0.2190\std{0.0032}
& 0.0565\std{0.0018} & 0.0555\std{0.0024} & 0.2614\std{0.0030}
& \underline{0.1563}\std{0.0038} & 0.0977\std{0.0012} & 0.5335\std{0.0041} & 0.2707\std{0.0004} \\
CoDiffMob
& 0.2106\std{0.0035} & 0.1176\std{0.0016} & 0.2494\std{0.0034}
& \underline{0.0422}\std{0.0009} & \underline{0.0420}\std{0.0011} & 0.2851\std{0.0042}
& 0.2180\std{0.0017} & \underline{0.0676}\std{0.0010} & \underline{0.5365}\std{0.0022} & \underline{0.2773}\std{0.0009} \\
LLMob
& \underline{0.1928}\std{0.0018} & \underline{0.1175}\std{0.0017} & \underline{0.1463}\std{0.0018}
& 0.0428\std{0.0005} & 0.0458\std{0.0009} & 0.2558\std{0.0037}
& 0.1920\std{0.0030} & 0.0723\std{0.0010} & 0.4995\std{0.0041} & 0.2596\std{0.0004} \\
LLM-COD
& 0.2374\std{0.0029} & 0.2232\std{0.0034} & 0.1976\std{0.0030}
& 0.0759\std{0.0015} & 0.0729\std{0.0015} & 0.2259\std{0.0034}
& 0.1919\std{0.0029} & 0.0754\std{0.0013} & 0.4684\std{0.0055} & 0.2682\std{0.0005} \\
TrajLLM
& 0.2364\std{0.0031} & 0.1880\std{0.0030} & 0.2207\std{0.0033}
& 0.0797\std{0.0016} & 0.0767\std{0.0017} & 0.2261\std{0.0032}
& 0.2084\std{0.0032} & 0.0939\std{0.0017} & 0.3750\std{0.0018} & 0.2389\std{0.0008} \\
AgentMove
& 0.2611\std{0.0037} & 0.2257\std{0.0038} & 0.2479\std{0.0035}
& 0.0436\std{0.0009} & 0.1753\std{0.0028} & \underline{0.2233}\std{0.0021}
& 0.2933\std{0.0045} & 0.0680\std{0.0009} & 0.5194\std{0.0029} & 0.2369\std{0.0040} \\
\rowcolor[gray]{0.92}
MOTION
& \textbf{0.1709}\std{0.0015} & \textbf{0.0958}\std{0.0014} & \textbf{0.1258}\std{0.0015}
& \textbf{0.0402}\std{0.0004} & \textbf{0.0403}\std{0.0007} & \textbf{0.2079}\std{0.0018}
& \textbf{0.1340}\std{0.0018} & \textbf{0.0629}\std{0.0007} & \textbf{0.5766}\std{0.0012} & \textbf{0.3190}\std{0.0005} \\
\midrule
\textit{Improvement}
& 11.36\% & 18.47\% & 14.01\%
& 4.74\% & 4.05\% & 6.90\%
& 14.27\% & 6.95\% & 7.47\% & 15.04\% \\
\midrule
\multicolumn{11}{c}{\textbf{Mobile Phone Dataset}} \\
\midrule
MoveSim
& 0.2382\std{0.0035} & 0.1920\std{0.0032} & 0.3151\std{0.0045}
& 0.1804\std{0.0030} & 0.1777\std{0.0031} & 0.3018\std{0.0047}
& 0.3387\std{0.0050} & 0.2231\std{0.0042} & 0.4366\std{0.0050} & 0.2253\std{0.0034} \\
DeepMobility
& \underline{0.1569}\std{0.0018} & \underline{0.1426}\std{0.0019} & 0.1996\std{0.0026}
& 0.1278\std{0.0014} & 0.1264\std{0.0056} & 0.3505\std{0.0059}
& 0.2892\std{0.0034} & 0.1855\std{0.0027} & 0.4550\std{0.0029} & 0.2622\std{0.0030} \\
CoDiffMob
& 0.2635\std{0.0039} & 0.1496\std{0.0027} & 0.1911\std{0.0025}
& 0.1165\std{0.0013} & 0.1163\std{0.0013} & 0.3478\std{0.0037}
& 0.2669\std{0.0051} & \underline{0.1320}\std{0.0041} & 0.4591\std{0.0067} & 0.2531\std{0.0025} \\
LLMob
& 0.1778\std{0.0029} & 0.1834\std{0.0030} & \underline{0.1116}\std{0.0018}
& 0.1277\std{0.0025} & 0.1274\std{0.0026} & 0.3544\std{0.0052}
& 0.2803\std{0.0043} & 0.2143\std{0.0045} & \underline{0.4601}\std{0.0062} & 0.3249\std{0.0032} \\
LLM-COD
& 0.1598\std{0.0026} & 0.1512\std{0.0025} & 0.2007\std{0.0031}
& 0.1139\std{0.0022} & 0.1134\std{0.0022} & 0.2982\std{0.0043}
& 0.2904\std{0.0045} & 0.1504\std{0.0032} & 0.4593\std{0.0041} & 0.2803\std{0.0036} \\
TrajLLM
& 0.1917\std{0.0031} & 0.1955\std{0.0033} & 0.1852\std{0.0030}
& \underline{0.1105}\std{0.0020} & \underline{0.1105}\std{0.0020} & \underline{0.2960}\std{0.0028}
& 0.2691\std{0.0038} & 0.2317\std{0.0048} & 0.4517\std{0.0049} & \underline{0.3574}\std{0.0049} \\
AgentMove
& 0.2257\std{0.0036} & 0.1443\std{0.0022} & 0.2818\std{0.0040}
& 0.1297\std{0.0026} & 0.1295\std{0.0027} & 0.3508\std{0.0051}
& \underline{0.2426}\std{0.0032} & 0.1921\std{0.0040} & 0.4367\std{0.0062} & 0.3270\std{0.0044} \\
\rowcolor[gray]{0.92}
MOTION
& \textbf{0.1367}\std{0.0020} & \textbf{0.1263}\std{0.0018} & \textbf{0.0955}\std{0.0014}
& \textbf{0.0925}\std{0.0015} & \textbf{0.0919}\std{0.0015} & \textbf{0.2773}\std{0.0022}
& \textbf{0.2150}\std{0.0024} & \textbf{0.1170}\std{0.0025} & \textbf{0.4971}\std{0.0025} & \textbf{0.4016}\std{0.0026} \\
\midrule
\textit{Improvement}
& 12.87\% & 11.43\% & 14.43\%
& 16.29\% & 16.83\% & 6.32\%
& 11.38\% & 11.36\% & 8.04\% & 12.37\% \\ 
\bottomrule
\end{tabular}%
}
\end{table*}

\subsection{Overall Performance}
Table~\ref{tab:overall_performance} reports the overall performance on the private car and mobile phone datasets. MOTION achieves the best results across all metrics on both datasets, demonstrating its effectiveness in preserving individual, population, and community-level mobility patterns.
At the \textbf{individual trajectory statistics} level, MOTION reduces Distance by 11.36\% over LLMob on the private car dataset and by 12.87\% over DeepMobility on the mobile phone dataset. These improvements suggest that community-conditioned priors and user-agent reasoning jointly support realistic individual movements.
At the \textbf{population and flow distributions} level, MOTION reduces Flow-EMD by 4.05\% over CoDiffMob on the private car dataset and by 16.83\% over TrajLLM on the mobile phone dataset, highlighting its ability to capture collective dynamics.
At the \textbf{community and graph-structural consistency} level, MOTION improves Motif, Comm-Size, and Comm-Mem by 14.27\%, 6.95\%, and 7.47\% on the private car dataset, and Motif and Comm-Mem by 11.38\% and 8.04\% on the mobile phone dataset, respectively. It also achieves the best Comm-Trans scores of 0.3190 and 0.4016, demonstrating superior community evolution consistency.
Overall, these results demonstrate that MOTION preserves community organization and community structure temporal evolution while maintaining realistic individual mobility patterns.

\begin{table}[t]
\centering
\caption{Ablation results on both datasets.}
\vspace{-3mm}
\label{tab:ablation}
\setlength{\tabcolsep}{0pt}
\renewcommand{\arraystretch}{1.08}
\footnotesize
\resizebox{0.92\columnwidth}{!}{%
\begin{tabular}{@{}>{\centering\arraybackslash}p{0.25\columnwidth}|*{5}{>{\centering\arraybackslash}p{0.12\columnwidth}}>{\centering\arraybackslash}p{0.15\columnwidth}@{}}
\toprule
\textbf{Variants} & \textbf{Dist.}$\downarrow$ & \textbf{Radius}$\downarrow$ & \textbf{P-EMD}$\downarrow$ & \textbf{C-Size}$\downarrow$ & \textbf{C-Mem}$\uparrow$ & \textbf{C-Trans.}$\uparrow$ \\
\midrule
\multicolumn{7}{c}{\textbf{Private Car Dataset}} \\
\midrule
W/o User & 0.1939 & 0.1918 & 0.0459 & 0.0887 & 0.4868 & 0.2790 \\
W/o Location & 0.1817 & 0.1050 & 0.0524 & 0.0881 & 0.5320 & 0.3103 \\
W/o Manager & 0.2054 & 0.1331 & 0.0486 & 0.0831 & 0.4677 & 0.2605 \\
W/o Reflection & 0.1949 & 0.1436 & 0.0556 & 0.1078 & 0.4855 & 0.2804 \\
W/o Community & 0.2178 & 0.1544 & 0.0616 & 0.0991 & 0.4615 & 0.2487 \\
Single-Agent & 0.2367 & 0.1174 & 0.0548 & 0.0792 & 0.5247 & 0.2539 \\
Deterministic & 0.2511 & 0.1219 & 0.0665 & 0.0841 & 0.4925 & 0.2042 \\
\rowcolor[gray]{0.92}
MOTION & \textbf{0.1709} & \textbf{0.0958} & \textbf{0.0402} & \textbf{0.0629} & \textbf{0.5766} & \textbf{0.3190} \\
\midrule
\multicolumn{7}{c}{\textbf{Mobile Phone Dataset}} \\
\midrule
W/o User & 0.1767 & 0.2367 & 0.1025 & 0.1608 & 0.4358 & 0.3205 \\
W/o Location & 0.1678 & 0.1674 & 0.1657 & 0.1975 & 0.4132 & 0.2972 \\
W/o Manager & 0.1863 & 0.1763 & 0.1327 & 0.1462 & 0.4682 & 0.2642 \\
W/o Reflection & 0.1889 & 0.1744 & 0.1967 & 0.1959 & 0.4173 & 0.2818 \\
W/o Community & 0.1927 & 0.1901 & 0.1419 & 0.2400 & 0.2997 & 0.2445 \\
Single-Agent & 0.1848 & 0.1769 & 0.1246 & 0.1735 & 0.4666 & 0.3331 \\
Deterministic & 0.2201 & 0.1930 & 0.1903 & 0.1368 & 0.4771 & 0.3187 \\
\rowcolor[gray]{0.92}
MOTION & \textbf{0.1367} & \textbf{0.1263} & \textbf{0.0925} & \textbf{0.1170} & \textbf{0.4971} & \textbf{0.4016} \\
\bottomrule
\end{tabular}%
}
\par\vspace{1mm}
\resizebox{0.95\columnwidth}{!}{\parbox{\columnwidth}{\scriptsize Dist.=Distance; P-EMD=Pop-EMD; C-Size/Mem/Trans=Comm-Size/Mem/Trans.}}
\vspace{-4mm}
\end{table}

\subsection{Ablation Analysis}
We conduct ablation studies to examine the contributions of community priors, agent roles, coordination, and reflection. \textbf{W/o User} and \textbf{W/o Location} remove the corresponding agents, while \textbf{W/o Manager} replaces manager coordination with fixed-weight proposal fusion and disables reflection. \textbf{W/o Community} removes community priors, while \textbf{W/o Reflection} retains manager coordination but disables reflection. \textbf{Single-Agent} uses the same priors, candidate sets, and reflection objectives as MOTION, with a comparable computational budget, but jointly considers user- and location-side evidence within a single agent without role separation. \textbf{Deterministic} retains the same information and reflection mechanism but replaces LLM-based reasoning with fixed-weight decision rules, whose parameters are optimized on a held-out validation subset of the training data.
As shown in Table~\ref{tab:ablation}, MOTION achieves the best results across all six metrics on both datasets. Removing the user agent degrades mobility and structural consistency, while removing the location agent yields the largest Pop-EMD among role-removal variants. Removing reflection reduces Comm-Trans from 0.1990 to 0.1604 and from 0.3819 to 0.2618 on the two datasets, while removing the manager further reduces it to 0.1405 and 0.2442, respectively. Removing community priors increases Comm-Size from 0.1170 to 0.2400 and reduces Comm-Mem from 0.4971 to 0.2997 and Comm-Trans from 0.3819 to 0.2245 on the mobile phone dataset. Moreover, MOTION outperforms both same-information controls, supporting the benefits of role-specific reasoning and coordination beyond shared priors. Appendix~\ref{appendix:reflection_baselines} further compares MOTION with baselines supplied with reflection information.

\begin{figure}[t]
    \centering
    \includegraphics[width=0.86\columnwidth]{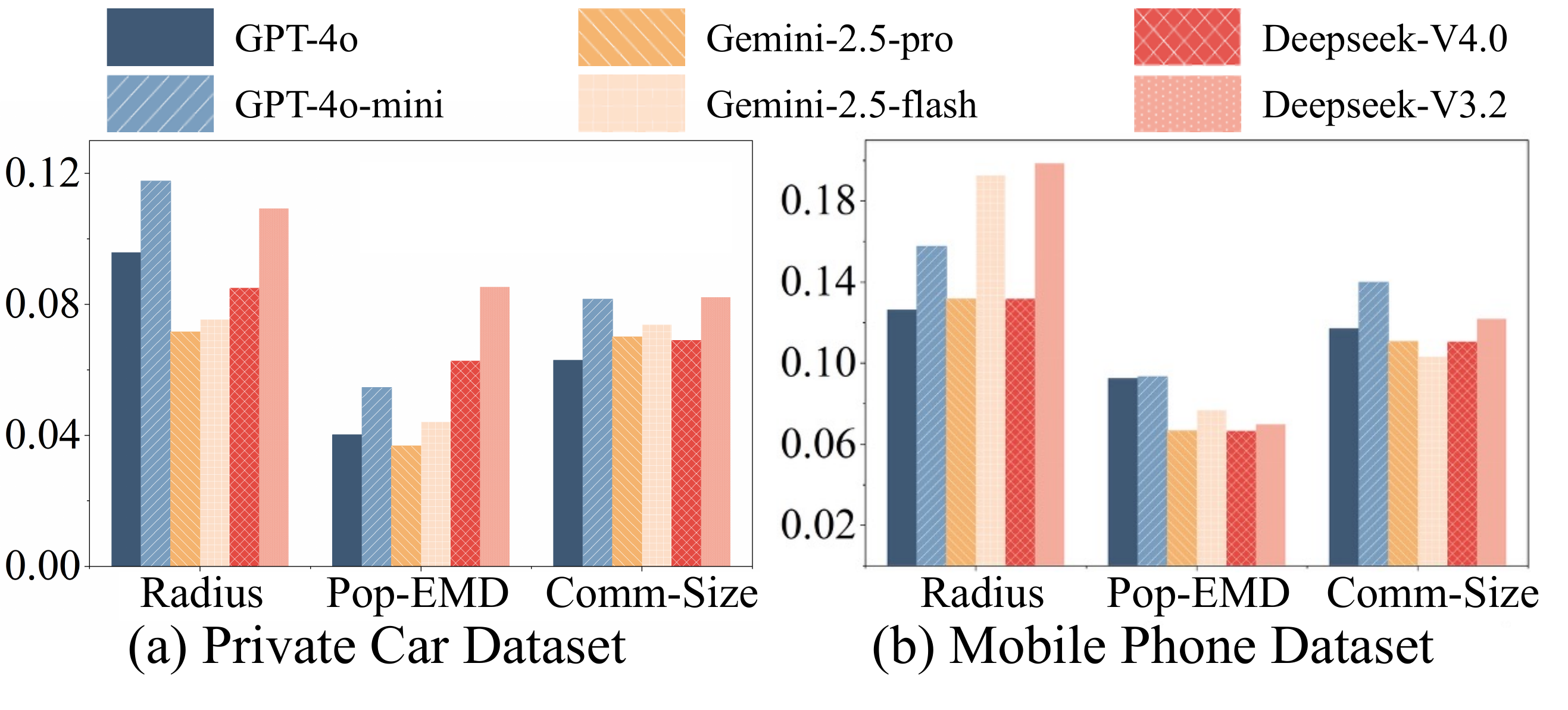}
    \vspace{-3mm}
    \caption{Impact of different LLM backbones on the private car and mobile phone datasets.}
    \label{fig:llm_backbone}
    \vspace{-5mm}
\end{figure}

\subsection{LLM Backbone Analysis}
We evaluate the impact of different LLM backbones by replacing the default GPT-4o with alternative models. As shown in Fig.~\ref{fig:llm_backbone}, backbone choice affects both individual mobility realism and community consistency, highlighting the importance of semantic understanding and coordination reasoning. MOTION remains effective across different backbones. We adopt GPT-4o for the main experiments based on its overall generation quality, although other models achieve better results on certain metrics.

\begin{figure*}[t]
    \centering
    \subfigure[Real]{
        \includegraphics[width=0.15\textwidth]{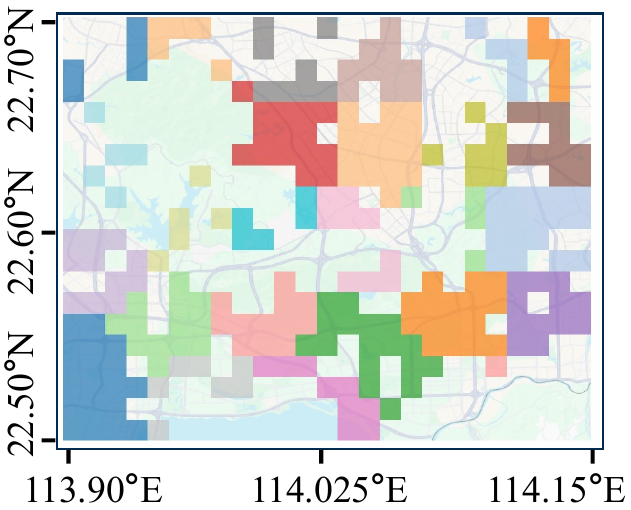}
    }\hfill
    \subfigure[CoDiffMob]{
        \includegraphics[width=0.15\textwidth]{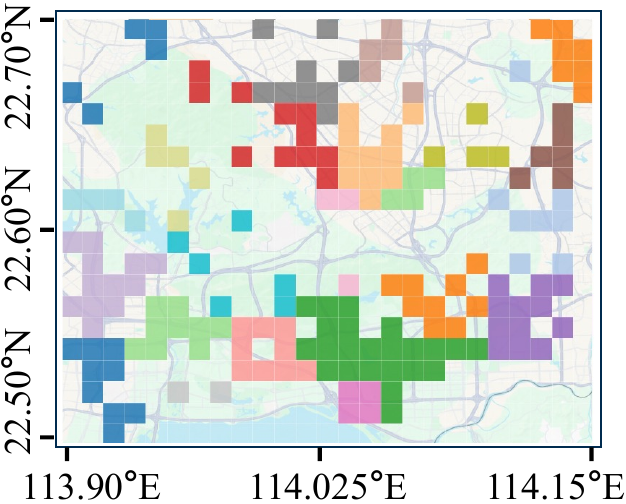}
    }\hfill
    \subfigure[MOTION]{
        \includegraphics[width=0.15\textwidth]{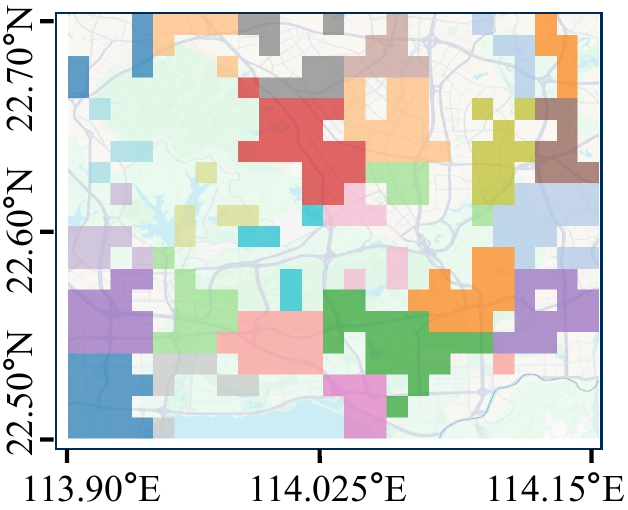}
    }\hfill
    \subfigure[Real]{
        \includegraphics[width=0.15\textwidth]{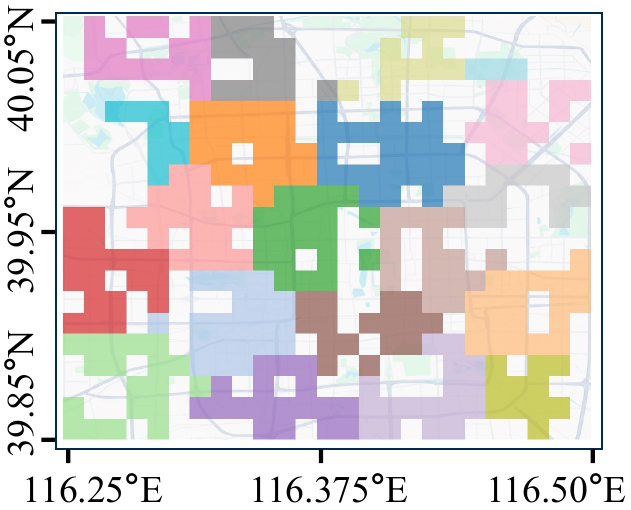}
    }\hfill
    \subfigure[LLMob]{
        \includegraphics[width=0.15\textwidth]{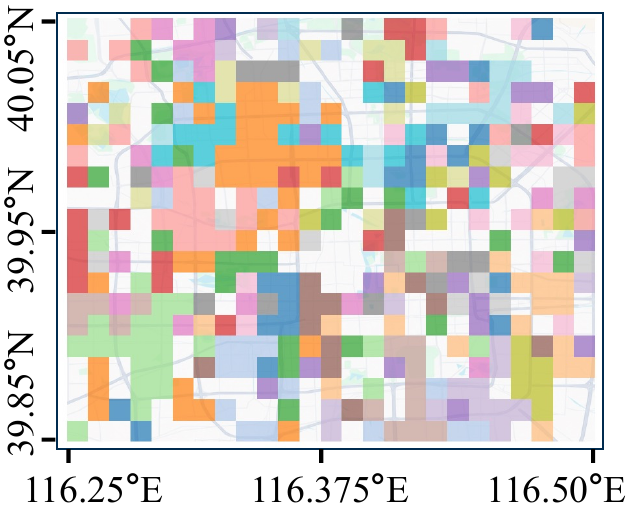}
    }\hfill
    \subfigure[MOTION]{
        \includegraphics[width=0.15\textwidth]{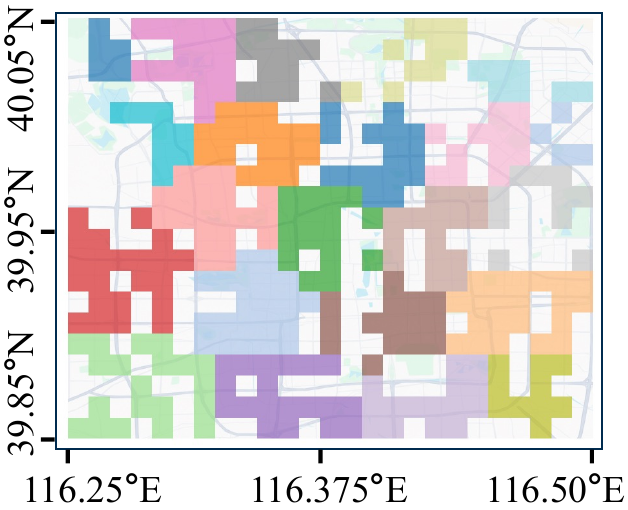}
    }
    \vspace{-4mm}
    \caption{Community structures detected from real and generated mobility.
    (a)--(c) Private car dataset; (d)--(f) Mobile phone dataset.
    Different colors represent different communities on the same gridded map.}
    \label{fig:community_visualization}
    \vspace{-2mm}
\end{figure*}
\vspace{-3mm}

\begin{figure}[t]
    \centering
    \includegraphics[width=0.8\columnwidth]{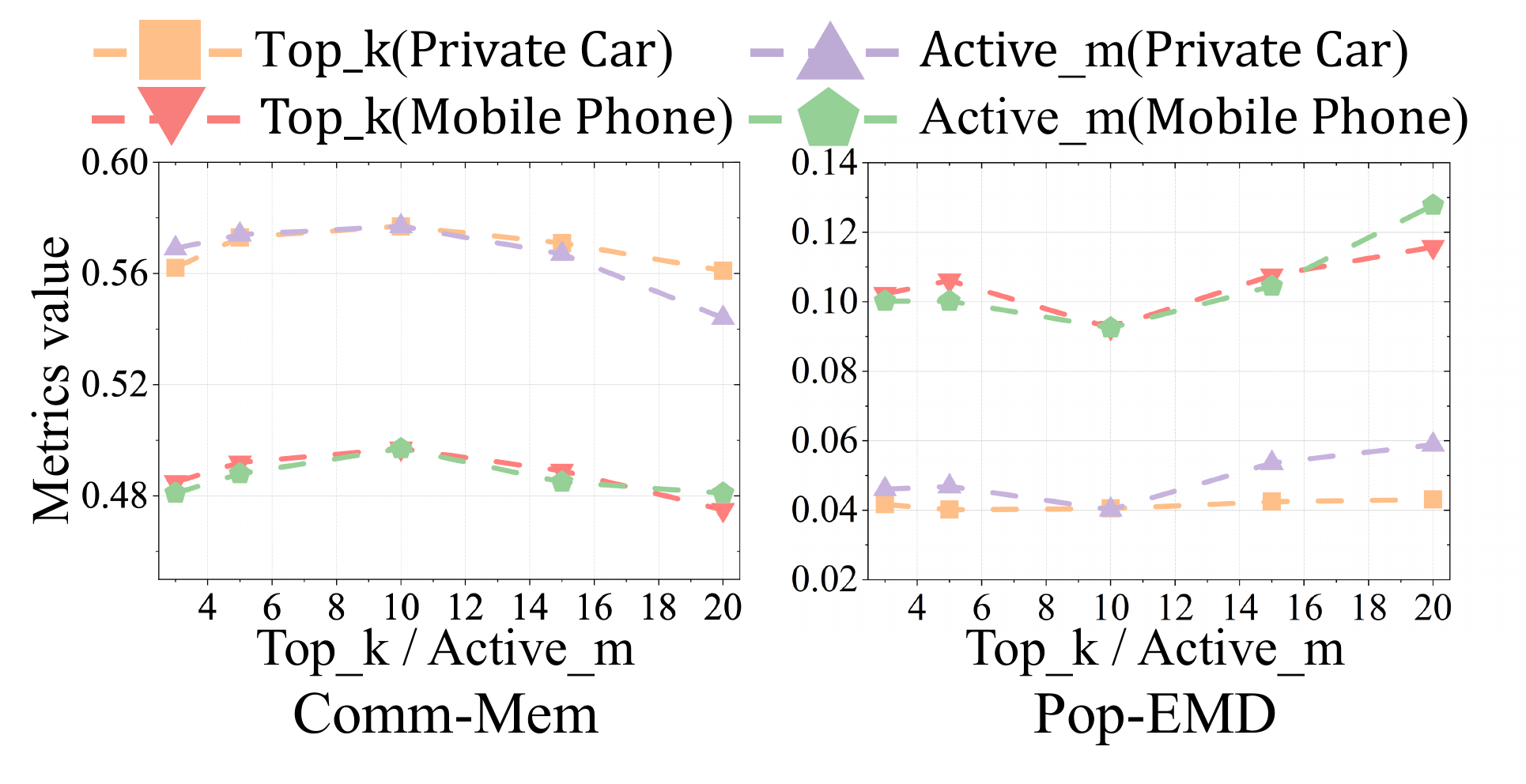}
    \vspace{-5mm}
    \caption{Sensitivity to \texttt{top\_k} and \texttt{active\_m} on the private car and mobile phone datasets.}
    \label{fig:hyperparameter}
    \vspace{-5mm}
\end{figure}

\subsection{Community Preservation Analysis}
To evaluate community structure preservation, we apply the same community detection procedure to real and generated user-location graphs and visualize the resulting communities in Fig.~\ref{fig:community_visualization}. CoDiffMob produces fragmented communities, while LLMob exhibits more mixed community assignments. In contrast, MOTION generates spatially coherent communities whose geographic extents and boundaries more closely resemble those of real mobility, demonstrating better preservation of mesoscopic community structures. Appendix~\ref{appendix:spatial_robustness} further shows that MOTION achieves the highest NMI under randomized, UL-only and UL-shuffled settings, suggesting robustness to backbone perturbations.


\subsection{Hyperparameter Analysis}
We analyze two key hyperparameters: \texttt{top\_k}, which determines the number of destination candidates retrieved for each user agent, and \texttt{active\_m}, which limits the number of locations sampled from each local set for agent activation. As shown in Fig.~\ref{fig:hyperparameter}, Comm-Mem and Pop-EMD generally improve as both parameters increase to 10 but deteriorate beyond this point, suggesting that excessive candidates introduce less relevant evidence and reasoning noise. These results highlight the importance of balancing evidence coverage with focused agent reasoning.

\subsection{Cross-Dataset Prior Recombination}
We evaluate cross-dataset prior recombination by transferring individual mobility motifs from a source dataset while using target-specific population, flow, co-mobility, and community priors to guide generation. This setting evaluates prior recombination rather than target-prior-free cross-city generalization. As shown in Fig.~\ref{fig:generalization} in Appendix~\ref{appendix:recombination}, Distance and Radius show larger discrepancies due to differences in mobility characteristics and spatial scales, while population-flow and community-level metrics remain relatively stable. These results suggest that target-specific priors help preserve structural consistency across datasets.

\begin{table}[t]
\centering
\caption{Performance on multiple held-out days (mean $\pm$ standard deviation). Bold indicates the best result.}
\vspace{-3mm}
\label{tab:temporal_heldout}
\setlength{\tabcolsep}{0pt}
\renewcommand{\arraystretch}{1.10}
\footnotesize
\resizebox{0.95\columnwidth}{!}{%
\begin{tabular}{@{}>{\centering\arraybackslash}m{\dimexpr0.2\columnwidth-\arrayrulewidth\relax}|*{4}{>{\centering\arraybackslash}m{0.2\columnwidth}}@{}}
\toprule
\rowcolor[gray]{0.92}
\textbf{Methods} & \textbf{Distance}$\downarrow$ & \textbf{Pop-EMD}$\downarrow$ & \textbf{Comm-Size}$\downarrow$ & \textbf{Comm-Mem}$\uparrow$ \\
\midrule
DeepMobility & $0.217\pm0.011$ & $0.053\pm0.004$ & $0.097\pm0.006$ & $0.536\pm0.004$ \\
CoDiffMob & $0.220\pm0.009$ & $0.043\pm0.003$ & $0.070\pm0.001$ & $0.549\pm0.003$ \\
LLMob & $0.277\pm0.013$ & $0.057\pm0.007$ & $0.143\pm0.002$ & $0.449\pm0.049$ \\
\rowcolor[gray]{0.92}
MOTION & $\mathbf{0.164}\pm0.006$ & $\mathbf{0.038}\pm0.002$ & $\mathbf{0.064}\pm0.003$ & $\mathbf{0.575}\pm0.010$ \\
\bottomrule
\end{tabular}%
}
\vspace{-2mm}
\end{table}

\begin{table}[t]
\centering
\caption{End-to-end token consumption and latency per generated user-day, and component costs per reasoning call.}
\vspace{-3mm}
\label{tab:complexity}
\setlength{\tabcolsep}{1.8pt}
\renewcommand{\arraystretch}{1.05}
\scriptsize
\resizebox{0.95\columnwidth}{!}{%
\begin{tabular}{@{}crrrr|crrrr@{}}
\toprule
\multicolumn{5}{c|}{\textbf{End-to-End / User-Day}} & \multicolumn{5}{c}{\textbf{MOTION Components / Call}} \\
\cmidrule(lr){1-5}\cmidrule(lr){6-10}
\multirow{2}{*}{\textbf{Method}} & \multicolumn{2}{c}{\textbf{Private Car}} & \multicolumn{2}{c|}{\textbf{Mobile Phone}}
& \multirow{2}{*}{\textbf{Component}} & \multicolumn{2}{c}{\textbf{Private Car}} & \multicolumn{2}{c}{\textbf{Mobile Phone}} \\
\cmidrule(lr){2-3}\cmidrule(lr){4-5}\cmidrule(lr){7-8}\cmidrule(lr){9-10}
& \textbf{Tokens} & \textbf{Lat.} & \textbf{Tokens} & \textbf{Lat.}
& & \textbf{Tokens} & \textbf{Lat.} & \textbf{Tokens} & \textbf{Lat.} \\
\midrule
TrajLLM & 43,804 & 231 & 44,185 & 230
& User & 436.1 & 1.57 & 372.3 & 2.21 \\
LLMob & \textbf{40,647} & 255 & 45,023 & 230
& Location & 1,345.0 & 2.15 & 1,462.0 & 1.98 \\
AgentMove & 42,331 & 264 & 48,159 & 244
& Coordination & 720.5 & 2.07 & 806.3 & 2.06 \\
MOTION & 40,662 & \textbf{229} & \textbf{43,132} & \textbf{224}
& Reflection & 866.4 & 3.77 & 981.0 & 3.11 \\
\bottomrule
\end{tabular}%
}
\par\vspace{0.5mm}
\resizebox{0.95\columnwidth}{!}{\parbox{\columnwidth}{\scriptsize Lat.=latency in seconds; User and Location denote agents.}}
\vspace{-4mm}
\end{table}

\subsection{Temporal Robustness Analysis}
We evaluate temporal robustness across multiple held-out days and longer generation horizons. As shown in Table~\ref{tab:temporal_heldout}, MOTION achieves the best results across Distance, Pop-EMD, Comm-Size, and Comm-Mem, demonstrating consistent performance across test days. As shown in Table~\ref{tab:temporal_continuous} in Appendix~\ref{appendix:temporal_robustness}, we further compare daily resetting with continuous generation over two to four days. Although daily resetting performs better, continuous generation maintains relatively stable population distributions and community consistency over longer horizons.

\subsection{Complexity Analysis}
Table~\ref{tab:complexity} reports user-day generation costs and component-level costs. MOTION achieves lower latency than LLMob and AgentMove with comparable or fewer tokens. We further conduct scaling experiments and find that costs scale approximately linearly with user count but remain relatively stable across location counts and reflection rounds. Active location selection costs $O(U_h+L_h\log L_h)$ per step, while local-set formation requires $O(d_{\mathrm{in}}+1)$ per active location. Potential optimizations include parallel execution and indexed retrieval.

\subsection{Case Study}
\begin{figure*}[t]
    \centering
    \includegraphics[width=0.75\textwidth]{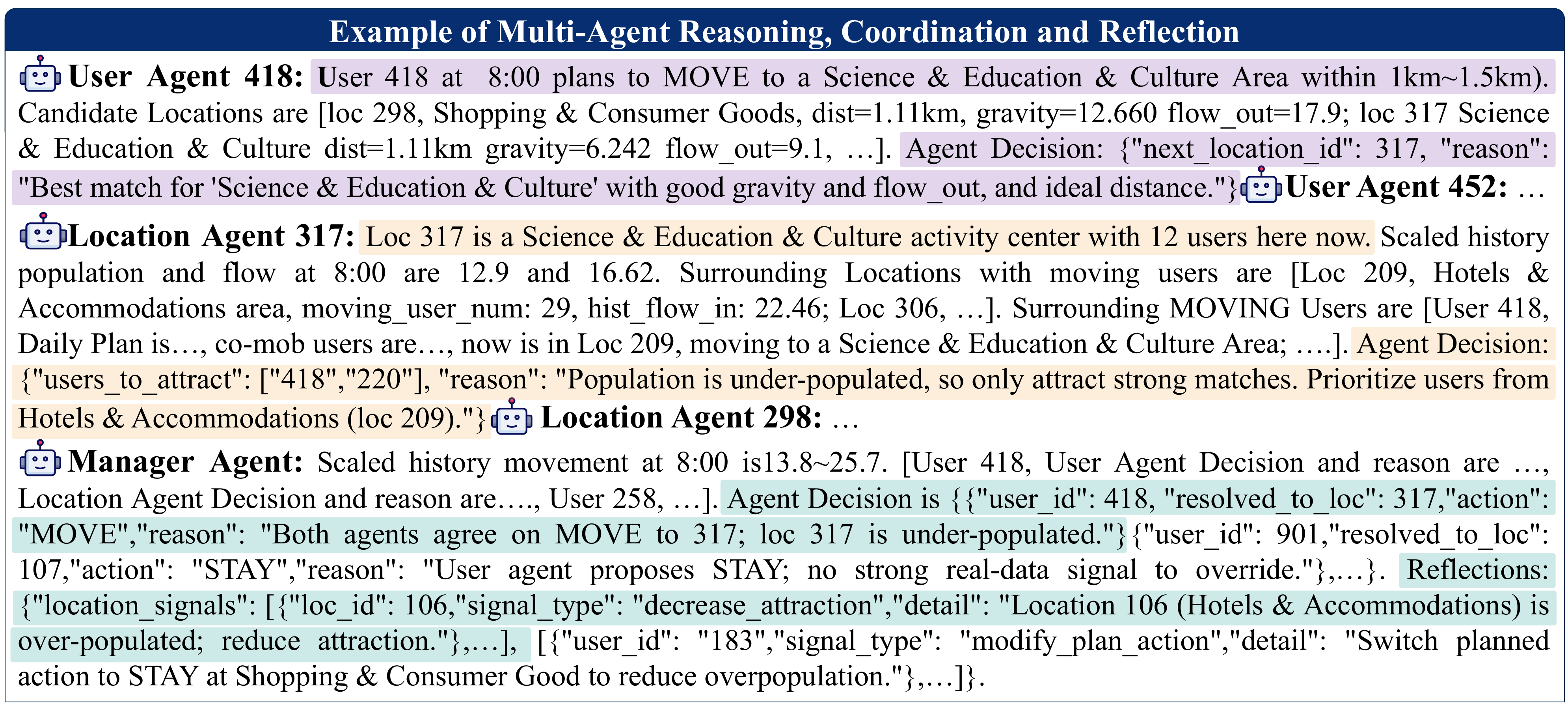}
    \vspace{-3mm}
    \caption{An example of multi-agent reasoning. Colored regions highlight the key inputs and outputs of each agent role.}
    \label{fig:case_study}
    \vspace{-3mm}
\end{figure*}

Fig.~\ref{fig:case_study} illustrates a graph update where user and location agents propose movements based on individual intentions and community priors, respectively. The manager coordinates both views, resolves conflicts, and provides targeted reflection signals to correct structural deviations. Detailed prompts are provided in Appendix~\ref{appendix:prompts}.

\begin{figure*}[t]
    \centering
    \includegraphics[width=0.9\textwidth]{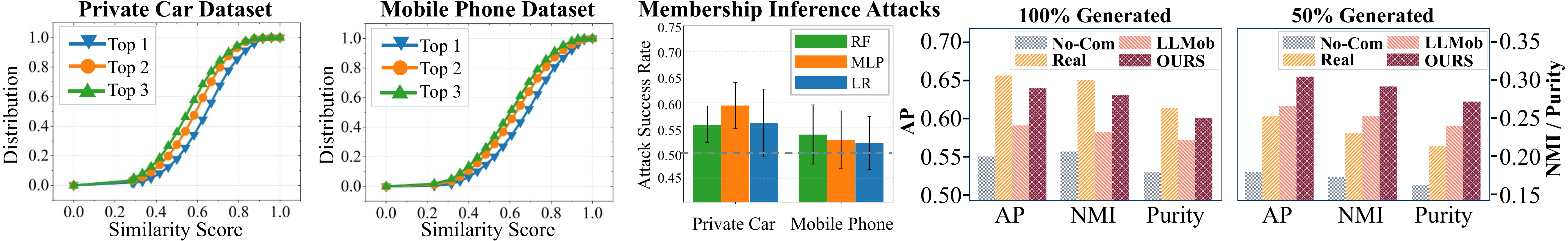}
    \vspace{-4mm}
    \caption{Uniqueness testing on two datasets (left). Membership inference attack accuracy (Middle). Downstream utility on private car dataset under 100\% and 50\% training settings. (Right).}
    \label{fig:privacy_and_downstream}
    \vspace{-3mm}
\end{figure*}

\vspace{-3mm}

\subsection{Privacy Analysis}
We assess privacy through uniqueness testing and membership inference attacks (MIA)~\cite{yuan2024generating,cao2025stage}. Using normalized Levenshtein similarity~\cite{levenshtein1966binary}, the left two panels of Fig.~\ref{fig:privacy_and_downstream} show moderate similarity between generated and training trajectories on both datasets, suggesting limited direct reproduction.
For MIA, the attacker uses four Levenshtein-based features without access to training records, model parameters, prompts, or identities. We evaluate 134 members and 134 non-members across five stratified, user-disjoint splits using RF~\cite{breiman2001random}, MLP~\cite{rosenblatt1958perceptron}, and LR~\cite{cox1958regression}. As shown in the middle panel of Fig.~\ref{fig:privacy_and_downstream}, attack accuracies are modestly above the 50\% random-guessing baseline. These results suggest no evidence of severe privacy exposure under the evaluated attack protocol.

\vspace{-2mm}

\subsection{Downstream Task Analysis}
We evaluate the utility of synthetic mobility for \textbf{Web-based location services} using CDGP~\cite{ji2023community},  focusing on future user-location interaction prediction (AP) and community recovery (NMI and Purity), which support personalized location services and community-aware user modeling. We compare CDGP trained on real data (\textbf{Real}), without community information (\textbf{No-Com}), and on data generated by \textbf{LLMob} or \textbf{MOTION}, using the same held-out real test set.
As shown in the right two panels of Fig.~\ref{fig:privacy_and_downstream}, MOTION outperforms LLMob and approaches Real when using 100\% generated training data. In the 50\% setting, LLMob and MOTION augment half the real data with an equal amount of synthetic data, while Real and No-Com use only the real half,demonstrating the utility of community-consistent mobility generation for Web-based user-location modeling and personalization.
\section{Related Work}
Early rule-based mobility generation method for Web-based location services, such as DRTISA~\cite{pappalardo2018data} and d-EPR~\cite{pappalardo2016human}, encode routines, exploration, and preferential return but rely on simplified behavioral assumptions.
Deep models such as SeqGAN~\cite{yu2017seqgan} and MoveSim~\cite{feng2020learning} learn user preferences and spatiotemporal transitions through adversarial generation.
Large language models (LLMs) provide semantic understanding and contextual reasoning for modeling user behaviors.
LLMob~\cite{wang2024large} verbalizes trajectory histories and contextual information for autoregressive next-location inference, whereas LLM-COD~\cite{yu2024harnessing} conditions destination generation on departure time, origin, and regional POI semantics.
LLM-based agents further incorporate memory, planning, tool use, and iterative feedback to simulate mobility decisions.
Ju et al.~\cite{ju2025trajllm} developed TrajLLM to hierarchically generate personas, activities, and destinations; Feng et al.~\cite{feng2025agentmove} introduced AgentMove, combining world knowledge and collective patterns for next-location reasoning.
Future research may leverage multi-agent collaboration to simulate user-location interactions and uncover deeper community structures underlying location-aware Web services~\cite{park2023generative,li2020evolvegraph}.
\section{Conclusion}
In this paper, we presented MOTION to support Web-based location services. MOTION extracts community-conditioned interaction priors, employs user agents and location agents to propose complementary graph updates, and uses a manager agent for coordination and reflection.
Experiments demonstrate improvements across individual, population, and community levels, while further analyses show cross-dataset prior recombination, privacy characteristics, and downstream utility for Web-based user-location modeling.

\clearpage
\begingroup
\bibliographystyle{ACM-Reference-Format}
\bibliography{Reference}
\endgroup
\clearpage

\newpage
\suppressfloats[t]
\section*{Supplementary Material}
\appendix

\begin{table*}[t]
\centering
\small
\vspace{-2mm}
\caption{Statistics of the private car and mobile phone datasets.}
\vspace{-2mm}
\label{tab:datasets}
\resizebox{0.90\textwidth}{!}{
\begin{tabular}{ccccccccc}
\toprule
\textbf{Dataset} & \textbf{Period} & \textbf{\# Locations} & \textbf{\# Users} & \textbf{\# POIs} & \begin{tabular}[c]{@{}c@{}}\textbf{\# POI}\\\textbf{Categories}\end{tabular} & \textbf{\# Records} & \begin{tabular}[c]{@{}c@{}}\textbf{Avg. Moves}\\\textbf{per User}\end{tabular} & \begin{tabular}[c]{@{}c@{}}\textbf{Avg. Daily Moves}\\\textbf{per User}\end{tabular} \\
\midrule
Private car dataset  & 2 weeks & 500 & 2,342 & 717,461 & 14 & 678,341 & 8.75 & 0.88 \\
Mobile phone dataset & 2 weeks & 500 & 5,352 & 679,413 & 14 & 1,235,619 & 18.60 & 1.28 \\
\bottomrule
\end{tabular}}
\vspace{-2mm}
\end{table*}

\section{Dataset Statistics and Ethical Statement}
\label{appendix:datasets}
The private car dataset consists of two weeks of GPS traces collected from 2,342 private vehicles.
The mobile phone dataset contains two weeks of GPS-based location traces collected from 5,352 mobile users.
To protect user privacy, coordinate-level locations are mapped to region-level identifiers, and each mobility record contains only a de-identified user or vehicle identifier, a timestamp, and the corresponding region identifier.
All data used in this study were collected with the explicit authorization of the individual users and data providers and were anonymized before we accessed them.
Table~\ref{tab:datasets} summarizes the resulting dataset statistics.

\begin{table}[t]
\centering
\caption{The 14 high-level POI semantic categories.}
\vspace{-2mm}
\label{tab:poi_categories}
\small
\setlength{\tabcolsep}{6pt}
\renewcommand{\arraystretch}{1.12}
\begin{tabular}{@{}p{0.46\columnwidth}p{0.46\columnwidth}@{}}
\toprule
\rowcolor[gray]{0.92}
\multicolumn{2}{c}{\textbf{POI Semantic Categories}} \\
\midrule
Transportation Facilities & Leisure \& Entertainment \\
\rowcolor[gray]{0.97}
Companies \& Enterprises & Healthcare \\
Commercial \& Residential & Tourist Attractions \\
\rowcolor[gray]{0.97}
Automotive & Life Services \\
Science \& Education \& Culture & Shopping \& Consumer Goods \\
\rowcolor[gray]{0.97}
Sports \& Fitness & Hotels \& Accommodations \\
Financial Institutions & Dining \& Cuisine \\
\bottomrule
\end{tabular}
\end{table}

\section{POI Profiles}
\label{appendix:poi}
To capture the functional semantics of each region, we construct a POI profile using an open-source POI database containing geographic coordinates, category labels, and place names.
As shown in Table~\ref{tab:poi_categories}, there are 14 top-level POI categories, while the underlying POI database contains finer-grained subcategories.
For each region, we compute the proportion of POIs belonging to each category, resulting in a 14-dimensional distribution vector that represents its functional semantics.
These regional POI profiles are associated with the mobility records and are used by MOTION for semantic mobility planning, destination retrieval, and location-side reasoning.

\section{Details of Motif Mining}
\label{appendix:prior_mining}
This section defines the mobility motifs extracted through metagraph-based prior mining and provides representative examples using POI semantics and discretized travel distances.
\textbf{STAY} represents a continuous period during which a user remains at the same location without an inter-location movement.
\textbf{MOVE\_AB} represents a one-way movement from a source location to a different destination without a return within the motif.
\textbf{ROUND\_ABA} describes a visit to another location followed by a return to the original source.
\textbf{CHAIN\_ABC} captures a sequence of visits to three distinct locations without revisiting an earlier location.
\textbf{RETURN\_ABCB} describes a movement through three locations followed by a return to the intermediate location rather than the original source.
\textbf{FULL\_ABCBA} captures a nested return pattern in which a user revisits the intermediate location and subsequently returns to the original source.
Each movement is represented by its source and destination POI semantics together with one of seven travel-distance bins: \texttt{0--1 km}, \texttt{1--2 km}, \texttt{2--3 km}, \texttt{3--5 km}, \texttt{5--8 km}, \texttt{8--12 km}, and \texttt{>12 km}.
Table~\ref{tab:motif_examples} provides concrete examples, where a repeated POI label within an example denotes a revisit to the same location instance; STAY has no distance bin because it contains no movement.

\begin{table*}[t]
\centering
\small
\caption{Representative mobility motif examples using POI semantics and discretized travel distances.}
\vspace{-2mm}
\label{tab:motif_examples}
\setlength{\tabcolsep}{5pt}
\begin{tabular}{p{0.14\textwidth}p{0.80\textwidth}}
\toprule
\textbf{Motif} & \textbf{Representative Examples} \\
\midrule
\texttt{STAY}
& Commercial \& Residential (no movement); Companies \& Enterprises (no movement). \\
\texttt{MOVE\_AB}
& Commercial \& Residential $\xrightarrow{\text{3--5 km}}$ Companies \& Enterprises;
Science \& Education \& Culture $\xrightarrow{\text{8--12 km}}$ Tourist Attractions. \\
\texttt{ROUND\_ABA}
& Commercial \& Residential $\xrightarrow{\text{2--3 km}}$ Shopping \& Consumer Goods
$\xrightarrow{\text{2--3 km}}$ Commercial \& Residential. \\
\texttt{CHAIN\_ABC}
& Commercial \& Residential $\xrightarrow{\text{1--2 km}}$ Transportation Facilities
$\xrightarrow{\text{5--8 km}}$ Companies \& Enterprises. \\
\texttt{RETURN\_ABCB}
& Hotels \& Accommodations $\xrightarrow{\text{0--1 km}}$ Dining \& Cuisine
$\xrightarrow{\text{3--5 km}}$ Leisure \& Entertainment
$\xrightarrow{\text{3--5 km}}$ Dining \& Cuisine. \\
\texttt{FULL\_ABCBA}
& Commercial \& Residential $\xrightarrow{\text{1--2 km}}$ Transportation Facilities
$\xrightarrow{\text{>12 km}}$ Tourist Attractions
$\xrightarrow{\text{>12 km}}$ Transportation Facilities
$\xrightarrow{\text{1--2 km}}$ Commercial \& Residential. \\
\bottomrule
\end{tabular}
\end{table*}

\section{Baseline Descriptions}
\label{appendix:baselines}
\textbf{MoveSim.}~\cite{feng2020learning}
MoveSim is a mobility-specific generative adversarial network that conditions trajectory generation on user identity and historical movements.
It learns latent representations of users and trajectories to capture semantic similarities among mobility behaviors, allowing generated sequences to reflect individual destination preferences and spatial transition patterns.
Domain-specific mobility regularities are incorporated into the adversarial generation process to improve trajectory realism.

\textbf{DeepMobility.}~\cite{yuan2025learning}
DeepMobility is a deep generative collaboration network that jointly models individual mobility behaviors and population-level flows. It employs a generative adversarial imitation learning framework with two collaborative mechanisms: bottom-up interaction modeling and top-down feedback refinement. The former incorporates population movement information into individual mobility decisions, while the latter uses aggregated flow patterns to guide the learning of individual behaviors. Through this bidirectional collaboration, DeepMobility generates individual trajectories that reproduce collective mobility dynamics.

\textbf{CoDiffMob.}~\cite{zhang2025noise}
CoDiffMob is a diffusion-based urban mobility generation model that incorporates collaborative noise priors to capture both individual movement characteristics and collective mobility patterns. It first constructs location transition sequences using population-level mobility flows and individual movement preferences. These sequences are then mapped into the noise space and combined with Gaussian noise to guide the diffusion process, enabling the generation of realistic individual trajectories while preserving collective flow patterns.

\textbf{LLMob.}~\cite{wang2024large}
LLMob formulates human mobility generation as an instruction-guided language modeling task.
It converts a user's historical trajectory and contextual information, such as time and location semantics, into textual prompts and autoregressively predicts subsequent movements.
This formulation exploits the sequential modeling and semantic reasoning capabilities of large language models to generate personalized mobility trajectories.

\textbf{LLM-COD.}
LLM-COD formulates destination prediction as a conditional language generation problem.
Its prompts encode the departure location, travel time, and regional point-of-interest context, enabling the language model to infer the semantic type of a plausible destination.
The predicted destination semantics are subsequently matched to candidate regions, allowing the model to use functional urban knowledge when generating origin--destination movements.

\textbf{TrajLLM.}~\cite{ju2025trajllm}
TrajLLM is a modular large-language-model-enhanced agent framework for daily trajectory simulation.
It organizes mobility generation into hierarchical stages that construct user personas, select activities, and determine corresponding destinations.
By incorporating demographic and behavioral context into these stages, TrajLLM generates interpretable daily routines that connect individual characteristics with activity and location choices.

\textbf{AgentMove.}~\cite{feng2025agentmove}
AgentMove is a large-language-model-based agentic framework for next-location prediction.
It decomposes mobility reasoning into a spatiotemporal memory module for individual patterns, a world-knowledge module for multi-scale urban structure, and a collective-knowledge module for shared population patterns.
The outputs of these modules are jointly interpreted by the language model to produce the final destination decision.

\section{Detailed Experimental Settings}
\label{appendix:settings}
We implement MOTION in Python and use LangGraph to orchestrate the three agent types.
Mobility records are discretized hourly over 500 POI-enriched regions.
We split users into training and test sets at a ratio of $8{:}2$, use the days preceding the final day to construct mobility priors, and evaluate 24-step generation from each test user's location at 00:00 on the final day.
We use GPT-4o with \texttt{top\_k}=10, \texttt{active\_m}=10, and at most 5 reflection rounds.
The LLM temperature is set to 0.2, and all experiments use a random seed of 42.
The same target users and initial locations are supplied to all methods whenever a baseline supports conditional initialization.

For dynamic graph construction, the spatial decay scale is set to 3 km, location edges longer than 10 km are removed, and the top 20\% of the remaining location-location edges are retained according to their combined spatial-flow weights.
For dynamic community detection, the Leiden resolution is set to 0.2; the initial graph is optimized for 20 iterations, subsequent snapshots are warm-started for 5 iterations, and the temporal smoothing coefficient is set to 0.4.

For destination retrieval, the population exponent is set to 1.0, the POI compatibility multiplier to 2.0, the minimum distance used by the gravity score to 0.1 km, and the long-distance exploration probability to 0.05.
Location agents prioritize locations with high incoming activity.
The co-mobility threshold is set to 0.6.
The manager agent compares intermediate graphs with priors extracted only from the training data and restricts each reflection round to affected users and locations within the current time step.

\section{Details of Evaluation Metrics}
\label{appendix:metrics}
Let $\mathcal{S}$ and $\widehat{\mathcal{S}}$ denote the real and generated trajectory sets, respectively, and let $d(\cdot,\cdot)$ denote geographic distance.
Following prior mobility-generation evaluations~\cite{feng2020learning,mauro2022generating}, we organize the metrics into individual trajectory statistics, population and flow distributions, and community and graph-structural consistency.
For distributional comparisons, we use the square root of the base-2 Jensen--Shannon divergence (JSD), defined using the base-2 Kullback--Leibler (KL) divergence as
\begin{equation}
\begin{aligned}
    \operatorname{JSD}(P,Q)
    &=\sqrt{\frac{1}{2}\operatorname{KL}_{2}(P\|M)
    +\frac{1}{2}\operatorname{KL}_{2}(Q\|M)},\\
    M&=\frac{P+Q}{2},
\end{aligned}
\end{equation}
where $\operatorname{JSD}(P,Q)\in[0,1]$.

\subsection{Individual Trajectory Statistics}
This level evaluates whether each generated trajectory preserves realistic spatial ranges and visitation diversity, using mobility statistics commonly adopted in trajectory generation~\cite{pappalardo2018data,feng2020learning}.

\textbf{Distance.}
Distance measures the average geographic displacement between consecutive time steps in a trajectory $s_i=[\ell_1^i,\ldots,\ell_T^i]$:
\begin{equation}
    D_i=\frac{1}{T-1}\sum_{t=2}^{T}d(\ell_{t-1}^i,\ell_t^i).
\end{equation}
We compute JSD between the empirical distributions of $D_i$ in $\mathcal{S}$ and $\widehat{\mathcal{S}}$.

\textbf{Radius.}
Radius measures the spatial extent of an individual's movements using the radius of gyration.
Let $\mathbf{x}_t^i$ be the coordinate of $\ell_t^i$ and $\bar{\mathbf{x}}_i=T^{-1}\sum_t\mathbf{x}_t^i$ be the trajectory center; then
\begin{equation}
    R_i=\sqrt{\frac{1}{T}\sum_{t=1}^{T}d(\mathbf{x}_t^i,\bar{\mathbf{x}}_i)^2}.
\end{equation}
Radius is the JSD between the real and generated distributions of $R_i$.

\textbf{Number of Locations (LocNum).}
LocNum measures the number of distinct locations visited by a user:
\begin{equation}
    L_i=\left|\{\ell_t^i\}_{t=1}^{T}\right|.
\end{equation}
We compare the distributions of $L_i$ using JSD to evaluate the diversity of visited locations.

\subsection{Population and Flow Distributions}
This level aggregates users and movements to evaluate whether generated mobility recovers the spatial organization of regional populations and origin--destination flows~\cite{rong2024interdisciplinary}.
We use Earth Mover's Distance (EMD)~\cite{rubner1998metric} because it accounts for the geographic cost of transporting probability mass between spatially different regions.

\textbf{Population Earth Mover's Distance (Pop-EMD).}
Pop-EMD evaluates whether users are distributed over geographically similar locations at each time step.
For two discrete spatial distributions $P$ and $Q$, Earth Mover's Distance (EMD) under geographic ground distance $d$ is
\begin{equation}
    \operatorname{EMD}_{d}(P,Q)
    =
    \min_{\pi\in\Pi(P,Q)}
    \sum_{x,y}\pi(x,y)d(x,y),
\end{equation}
where $\Pi(P,Q)$ denotes the set of transport plans whose marginals are $P$ and $Q$.
Let $P_t^{\mathrm{pop}}$ and $\widehat{P}_t^{\mathrm{pop}}$ denote the real and generated population distributions at time $t$; Pop-EMD is
\begin{equation}
    \operatorname{Pop\text{-}EMD}
    =\frac{1}{T}\sum_{t=1}^{T}
    \operatorname{EMD}_{d}(P_t^{\mathrm{pop}},\widehat{P}_t^{\mathrm{pop}}).
\end{equation}

\textbf{Flow Earth Mover's Distance (Flow-EMD).}
Flow-EMD measures whether generated origin--destination movements reproduce the spatial organization of real population flows.
Each mobility event is represented by $e=(o,d)$, including a stay event when $o=d$, and the distance between two events is
\begin{equation}
    d_{F}(e,e')=\frac{d(o,o')+d(d,d')}{2}.
\end{equation}
Let $P_t^{\mathrm{flow}}$ and $\widehat{P}_t^{\mathrm{flow}}$ denote the corresponding event distributions; Flow-EMD is
\begin{equation}
    \operatorname{Flow\text{-}EMD}
    =\frac{1}{T}\sum_{t=1}^{T}
    \operatorname{EMD}_{d_F}(P_t^{\mathrm{flow}},\widehat{P}_t^{\mathrm{flow}}).
\end{equation}

\textbf{G-Rank.}
G-Rank evaluates whether generated mobility preserves the popularity distribution of major locations.
Let $\mathcal{L}_{100}$ be the 100 most visited locations in the real data, and let $n_\ell$ and $\widehat n_\ell$ denote their real and generated visit counts.
Their normalized distributions are
\begin{equation}
    p_\ell=\frac{n_\ell}{\sum_{j\in\mathcal{L}_{100}}n_j},
    \qquad
    \widehat p_\ell=\frac{\widehat n_\ell}{\sum_{j\in\mathcal{L}_{100}}\widehat n_j}.
\end{equation}
G-Rank is defined as $\operatorname{JSD}(\mathbf{p},\widehat{\mathbf{p}})$ over the shared set $\mathcal{L}_{100}$.

\subsection{Community and Graph-Structural Consistency}
This level evaluates whether the generated graph sequence preserves the mesoscopic community organization and higher-order interaction structures emerging across users~\cite{bassolas2019hierarchical,ji2023community}.
It complements trajectory and aggregate-flow metrics with recurring mobility structures, community size and membership consistency, and pairwise community transitions over time~\cite{bojchevski2018netgan,mauro2022generating}.

\textbf{Motif.} We compute JSD between normalized occurrence distributions of the five non-stay motifs in Table~\ref{tab:motif_examples}.

\textbf{Comm-Size.} Comm-Size measures community size consistency across sliding windows. For each window $w\in\mathcal W$, Leiden detects communities from aggregated user-location graphs. Given partition $\mathcal C_w$, the size distribution is:
\begin{equation}
p_w^C(b)=\frac{1}{|\mathcal C_w|}\sum_{C\in\mathcal C_w}
\mathbb I\!\left[\frac{|C|}{|\mathcal V_w|}\in b\right],\quad b\in\mathcal B,
\end{equation}
where $\mathcal V_w$ is the node set and $\mathcal B$ denotes shared bins. We compute:
\begin{equation}
\operatorname{Comm\text{-}Size}=
\frac{1}{|\mathcal W|}\sum_{w\in\mathcal W}
\operatorname{JSD}(p_w^C,\widehat p_w^C).
\end{equation}

\textbf{Comm-Mem.} Comm-Mem evaluates community membership consistency. We aggregate visits over 23 hourly steps into bipartite graphs with unit-weight user-location edges and no static backbone. Leiden with modularity (resolution 1.0) independently detects communities. Let $A_k^L$ and $B_k^L$ denote shared location labels from real and generated graphs under seeds $k$ and $100+k$. We compute:
\begin{equation}
\operatorname{Comm\text{-}Mem}=
\frac{1}{5}\sum_{k=0}^{4}
\frac{2I(A_k^L;B_k^L)}
{H(A_k^L)+H(B_k^L)},
\end{equation}
where $I$ and $H$ denote mutual information and entropy. Higher Comm-Mem indicates better membership consistency.

\textbf{Comm-Trans.} Comm-Trans evaluates community evolution using overlapping 8-hour windows with a 1-hour stride. Leiden detects communities from aggregated user-location graphs without static edges. Let $c_w(\ell_i)$ denote the community label of location $\ell_i$ in window $w$. For a location pair, define:
\begin{equation}
\begin{aligned}
S_w(i,j)&=\mathbb I[c_w(\ell_i)=c_w(\ell_j)],\\
\Delta_w(i,j)&=S_{w+1}(i,j)-S_w(i,j),
\end{aligned}
\end{equation}
where $\Delta_w=-1,0,1$ denotes separation, no change, or convergence, respectively. Let $\mathcal T^R$ and $\mathcal T^G$ denote real and generated nonzero transitions $(w,i,j,\Delta_w(i,j))$. We compute:
\begin{equation}
\operatorname{Comm\text{-}Trans}
=\frac{2|\mathcal T^R\cap\mathcal T^G|}
{|\mathcal T^R|+|\mathcal T^G|}.
\end{equation}
This transition F1 score lies in $[0,1]$ (higher is better) and matches both the timing and direction of pairwise community changes. Co-membership comparisons are invariant to community-label permutations; unchanged pairs are excluded.

\section{Additional Privacy Results}
\label{appendix:privacy}
\begin{figure}[t]
    \centering
    \includegraphics[width=\columnwidth]{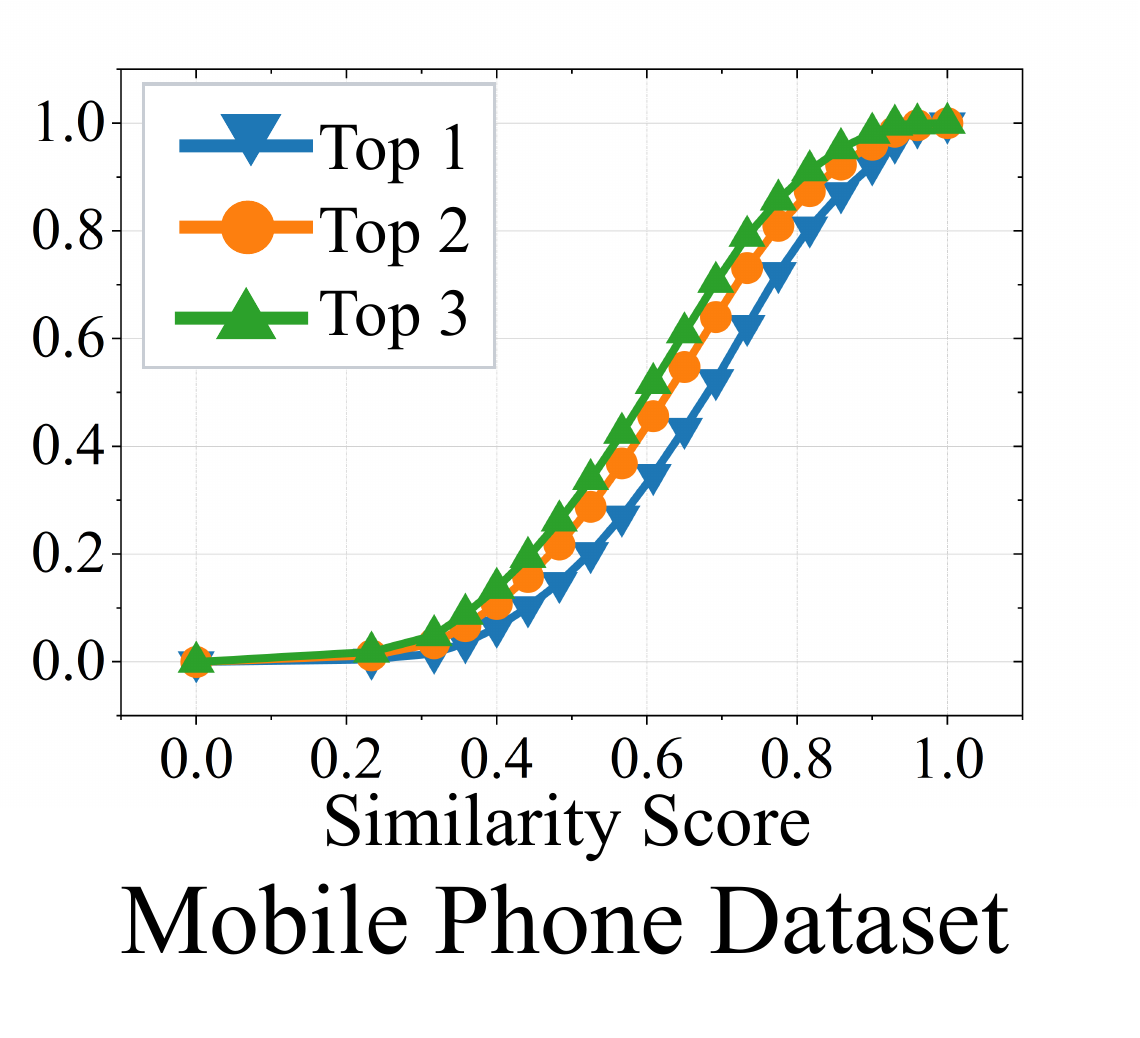}
    \caption{Cumulative distributions of the Top-1, Top-2, and Top-3 similarities between generated trajectories and their nearest real trajectories in the training split of the mobile phone dataset.}
    \label{fig:uniqueness_bj}
\end{figure}

To complement the private car dataset results in the main text, we repeat the uniqueness test on the mobile phone dataset.
For each generated 24-hour trajectory, we compute its normalized Levenshtein similarity~\cite{levenshtein1966binary} to every real trajectory in the training split and retain the Top-1, Top-2, and Top-3 nearest matches, following prior privacy evaluations of generated mobility~\cite{yuan2024generating,cao2025stage}.
Fig.~\ref{fig:uniqueness_bj} reports the cumulative distributions of these similarities.
The Top-1 similarities are concentrated in an intermediate range, showing that generated trajectories retain recognizable daily mobility regularities, while the Top-2 and Top-3 curves shift toward lower similarity.
Only a small tail approaches exact matching, and similarity consistently decreases as more neighbors are considered.
This pattern agrees with the results on the private car dataset and suggests that MOTION does not repeatedly reproduce a small set of training trajectories.
Together with the membership inference results in the main text, the analysis provides empirical evidence that MOTION learns reusable mobility patterns while limiting sample-specific memorization.

\section{Robustness to the Location Backbone}
\label{appendix:spatial_robustness}
To examine whether community consistency depends on static backbone information or specific user-location interactions, we evaluate three perturbations: randomizing location-location flow weights (\textbf{Randomized}), removing location-location edges (\textbf{UL-only}), and degree-preserving random rewiring of user-location edges (\textbf{UL-shuffled}). We aggregate interactions over 23 hourly steps for community detection. As shown in Table~\ref{tab:backbone_robustness}, MOTION maintains its advantage without the original spatial backbone. The UL-shuffled control further examines whether community agreement reflects co-visitation structure beyond node-level visit statistics.

\begin{table}[!htbp]
\centering
\caption{Community consistency under backbone perturbations (NMI$\uparrow$).}
\label{tab:backbone_robustness}
\scriptsize
\setlength{\tabcolsep}{2pt}
\renewcommand{\arraystretch}{1.10}
\resizebox{0.95\columnwidth}{!}{%
\begin{tabular}{@{}l|cccc>{\columncolor[gray]{0.92}}c@{}}
\toprule
\textbf{Setting} & \textbf{CoDiffMob} & \textbf{DeepMobility} & \textbf{TrajLLM} & \textbf{LLMob} & \textbf{MOTION} \\
\midrule
Randomized & $0.73\pm0.04$ & $0.73\pm0.03$ & $0.70\pm0.01$ & $0.65\pm0.03$ & $\mathbf{0.76}\pm0.02$ \\
UL-only & $0.55\pm0.03$ & $0.54\pm0.02$ & $0.51\pm0.01$ & $0.40\pm0.02$ & $\mathbf{0.58}\pm0.02$ \\
UL-shuffled & $0.47\pm0.03$ & $0.48\pm0.02$ & $0.45\pm0.04$ & $0.33\pm0.02$ & $\mathbf{0.50}\pm0.03$ \\
\bottomrule
\end{tabular}}
\end{table}

\section{Reflection Analysis}
\label{appendix:reflection_rounds}
\begin{figure}[t]
    \centering
    \includegraphics[width=\columnwidth]{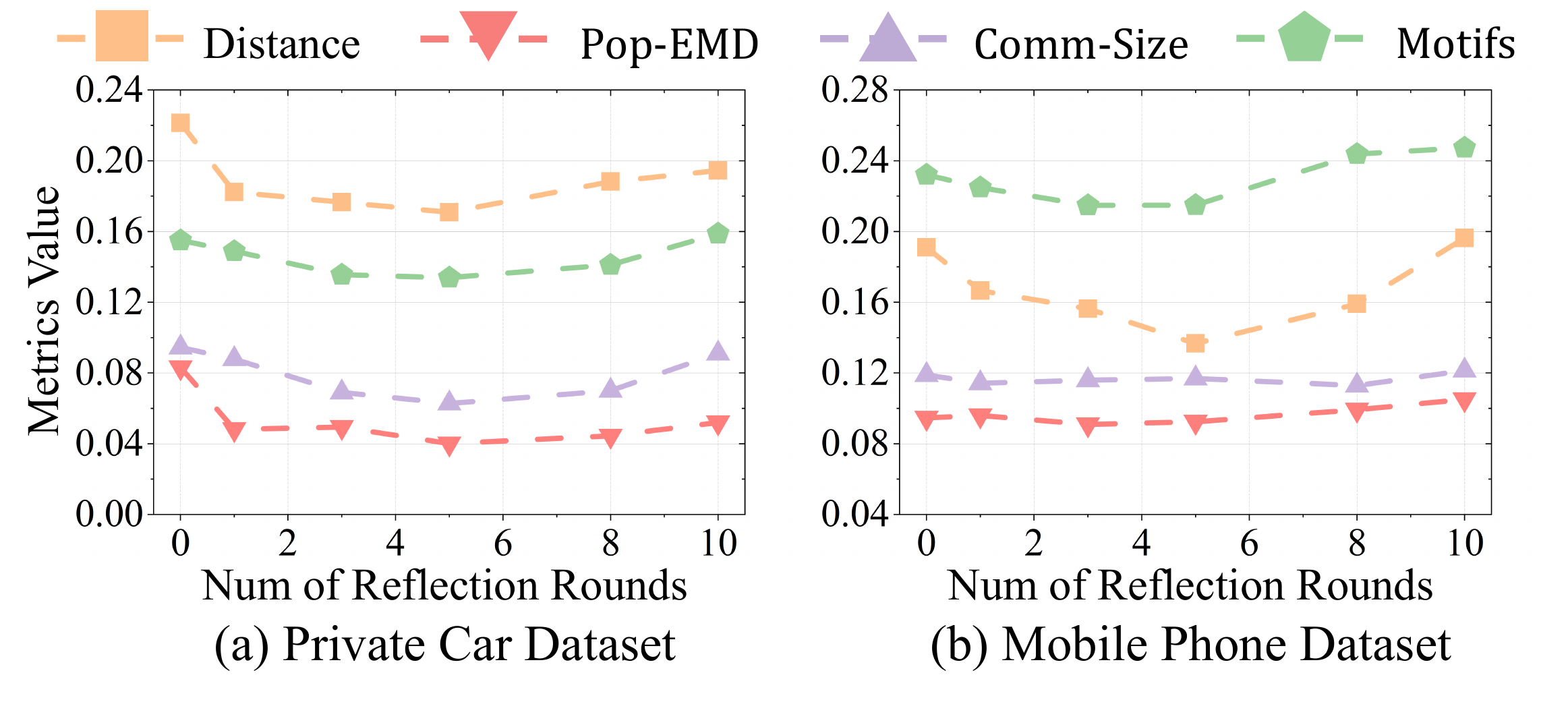}
    \caption{Impact of the maximum number of within-step reflection rounds on the private car and mobile phone datasets. Lower values indicate better performance.}
    \label{fig:reflection_round}
\end{figure}

Fig.~\ref{fig:reflection_round} shows that performance generally improves over the first five reflection rounds, with inconsistent gains thereafter; we therefore use five rounds.

We also add reflection information to CoDiffMob and LLM-COD (``+'') on the private car dataset.
MOTION leads all six metrics in Table~\ref{tab:reflection_baselines}, including Comm-Trans (0.1990 versus 0.1524 for CoDiffMob+ and 0.1511 for LLM-COD+), suggesting benefits from coordinated multi-agent reasoning beyond access to reflection information.

\begin{table}[!htbp]
\centering
\caption{Comparison with baselines supplied with reflection information (``+'') on the private car dataset. Bold indicates the best result.}
\label{tab:reflection_baselines}
\scriptsize
\setlength{\tabcolsep}{2pt}
\renewcommand{\arraystretch}{1.10}
\resizebox{0.95\columnwidth}{!}{%
\begin{tabular}{@{}l|rrrrrr@{}}
\toprule
\textbf{Method} & \textbf{Dist.}$\downarrow$ & \textbf{LocNum}$\downarrow$ & \textbf{P-EMD}$\downarrow$ & \textbf{C-Size}$\downarrow$ & \textbf{C-Mem}$\uparrow$ & \textbf{C-Trans.}$\uparrow$ \\
\midrule
CoDiffMob+ & 0.2097 & 0.2523 & 0.0422 & 0.0663 & 0.5490 & 0.1524 \\
LLM-COD+ & 0.2173 & 0.1963 & 0.0486 & 0.0736 & 0.5461 & 0.1511 \\
\rowcolor[gray]{0.92}
MOTION & \textbf{0.1709} & \textbf{0.1258} & \textbf{0.0402} & \textbf{0.0629} & \textbf{0.5766} & \textbf{0.1990} \\
\bottomrule
\end{tabular}}
\par\smallskip
\parbox{0.95\columnwidth}{\scriptsize Dist.=Distance; P-EMD=Pop-EMD; C-Size/Mem/Trans=Comm-Size/Mem/Trans.}
\end{table}

\section{Additional Generalization Results}
\label{appendix:generalization}
\label{appendix:recombination}
\begin{figure}[t]
    \centering
    \includegraphics[width=\columnwidth]{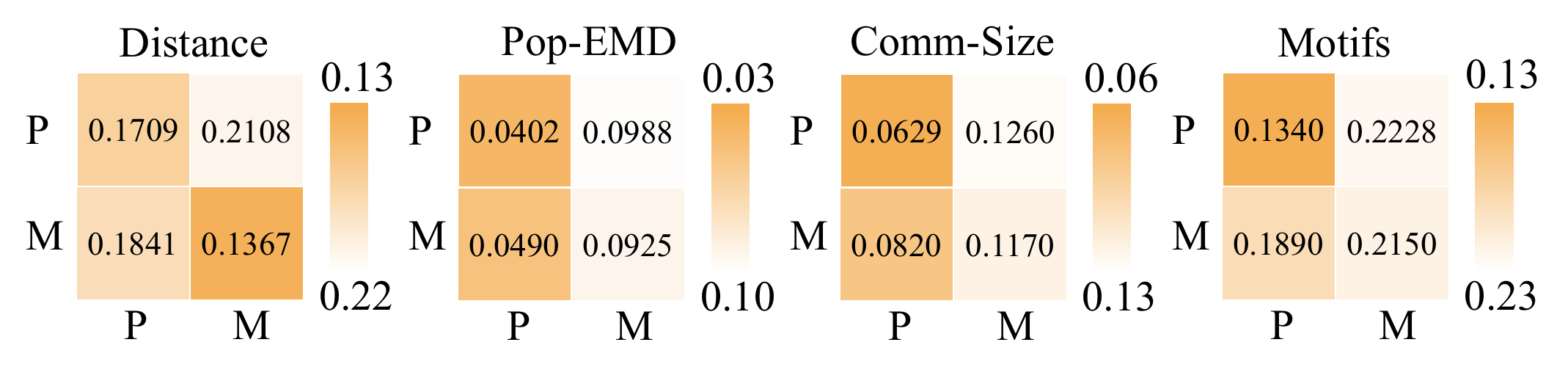}
    \caption{Cross-dataset generalization between the private car dataset (P) and mobile phone dataset (M). Rows indicate the source dataset, while columns indicate the target dataset.}
    \label{fig:generalization}
\end{figure}

Fig.~\ref{fig:generalization} reports cross-dataset generalization when user mobility motifs are transferred from a source dataset and interaction priors are extracted from the target dataset.
Rows indicate the source of the mobility motifs used to construct daily plans, whereas columns indicate the target dataset from which the interaction priors are extracted and on which generation is evaluated.
The diagonal cells correspond to within-dataset generation, and the off-diagonal cells evaluate recombination across datasets.
Distance and Radius show larger cross-dataset gaps, particularly when private-car motifs are transferred to mobile-phone mobility, because the two datasets capture different travel modes, activity ranges, and spatial scales.
In contrast, Pop-EMD, Flow-EMD, Comm-Size, and Motif remain substantially more stable.
For example, when transferring private-car motifs to the mobile phone dataset, Flow-EMD changes only from 0.0919 to 0.0944 and Comm-Size from 0.1170 to 0.1260.
These results show that target-specific interaction priors can anchor population-flow organization and community structure even when the transferred individual mobility scale differs from that of the target dataset.

\section{Additional Temporal Robustness Results}
\label{appendix:temporal_robustness}
We compare multi-day generation with daily resetting (Reset) and uninterrupted generation (No Reset) over two to four days.
Table~\ref{tab:temporal_continuous} reports Pop-EMD and Comm-Mem at each horizon.
Daily resetting yields lower Pop-EMD and higher Comm-Mem at every horizon.
Without daily resetting, Pop-EMD stays between 0.0444 and 0.0452, while Comm-Mem declines from 0.5473 to 0.5304, indicating gradual community-membership drift over longer horizons.

\begin{table}[t]
\centering
\caption{Multi-day generation with daily resetting (Reset) or uninterrupted generation (No Reset). Bold marks the better result for each metric and horizon.}
\label{tab:temporal_continuous}
\setlength{\tabcolsep}{0pt}
\renewcommand{\arraystretch}{1.10}
\footnotesize
\begin{tabular}{@{}>{\centering\arraybackslash}m{\dimexpr0.12\columnwidth-2\arrayrulewidth\relax}|*{2}{>{\centering\arraybackslash}m{0.22\columnwidth}}|*{2}{>{\centering\arraybackslash}m{0.22\columnwidth}}@{}}
\toprule
\multirow{2}{*}{\textbf{Days}} & \multicolumn{2}{c|}{\textbf{Pop-EMD}$\downarrow$} & \multicolumn{2}{c}{\textbf{Comm-Mem}$\uparrow$} \\
\cmidrule(lr){2-3}\cmidrule(lr){4-5}
& \textbf{Reset} & \textbf{No Reset} & \textbf{Reset} & \textbf{No Reset} \\
\midrule
2 & \textbf{0.0417} & 0.0451 & \textbf{0.5642} & 0.5473 \\
3 & \textbf{0.0415} & 0.0444 & \textbf{0.5779} & 0.5371 \\
4 & \textbf{0.0348} & 0.0452 & \textbf{0.5803} & 0.5304 \\
\bottomrule
\end{tabular}
\end{table}

\section{Component-Level Complexity Analysis}
\label{appendix:component_complexity}
The right panel of Table~\ref{tab:complexity} reports the average token consumption and latency of each reasoning component per call under the same LLM backbone and experimental configuration.
The user agent has the lightest context because it selects among a compact destination set conditioned on one user's mobility plan, requiring 372--436 tokens and 1.57--2.21 seconds per call.
The location agent consumes the most tokens because it jointly reasons over regional semantics, expected population and flow states, nearby users, and co-mobility evidence.
Coordination uses a moderate context containing only the conflicting individual- and community-view proposals and therefore maintains a latency of approximately two seconds on both datasets.
Reflection has the highest latency despite using fewer tokens than location reasoning because it diagnoses several node and edge discrepancies and converts them into structured, agent-specific correction signals.
The component costs are similar across the two datasets, indicating that the reasoning burden is determined mainly by agent roles rather than a particular mobility source.
These per-call costs do not directly sum to the end-to-end user-day cost: active location selection limits location-agent invocations, and targeted reflection reactivates only agents associated with detected discrepancies.
This selective execution keeps the overall complexity comparable to user-centric LLM generators, as reported in the main text.

\section{Additional Hyperparameter Results}
\label{appendix:hyperparameters}
\begin{figure}[t]
    \centering
    \includegraphics[width=\columnwidth]{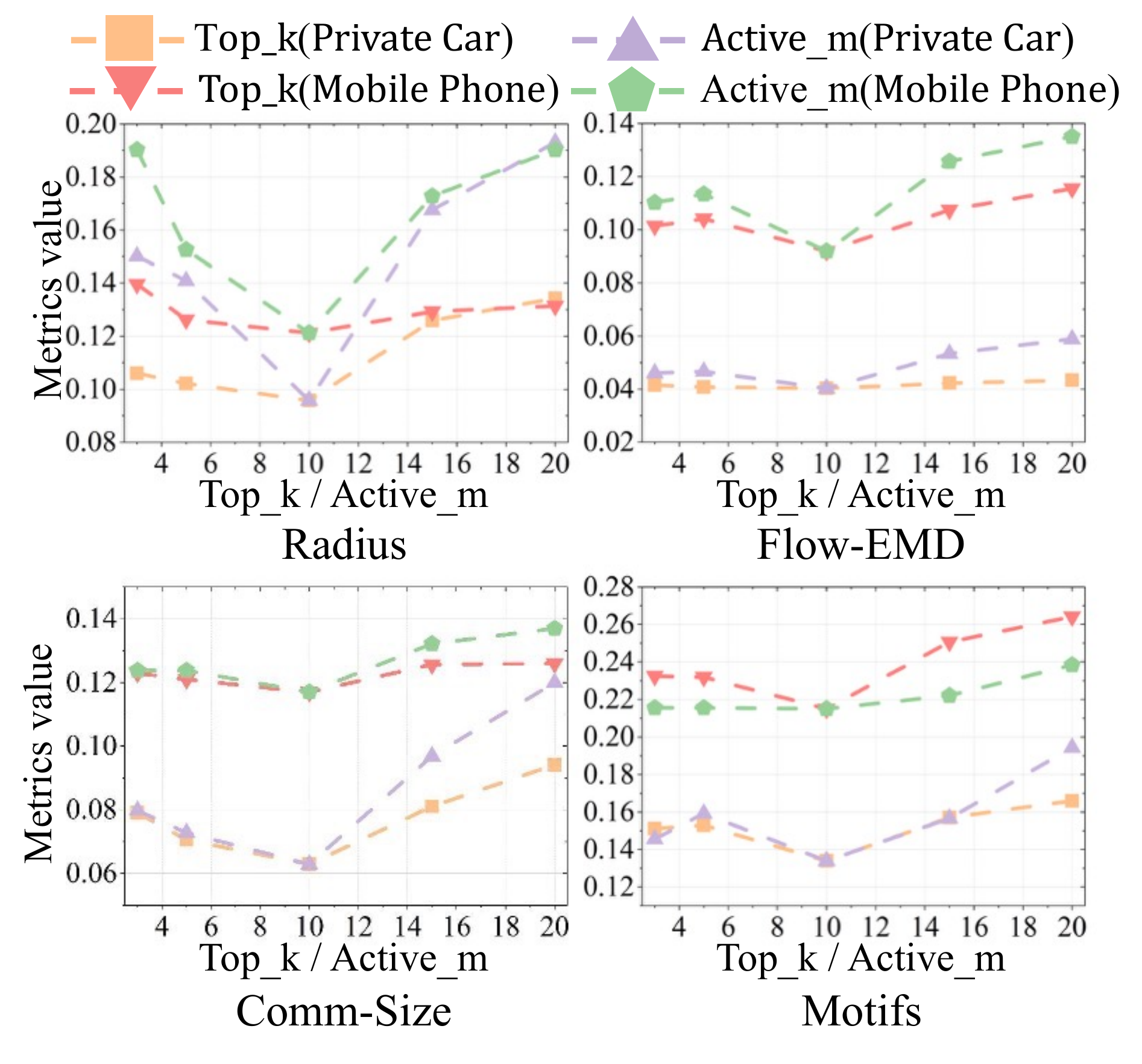}
    \caption{Additional sensitivity results of \texttt{top\_k} and \texttt{active\_m} on the private car and mobile phone datasets. Lower values indicate better performance.}
    \label{fig:hyper_appendix}
\end{figure}

We further evaluate \texttt{top\_k} and \texttt{active\_m} using Radius, Flow-EMD, Comm-Size, and Motif, complementing the Comm-Mem and Pop-EMD results in the main text.
The former controls how many destination candidates are provided to each user agent, while the latter limits the number of locations sampled from each structurally relevant local set for location-agent activation.
We vary one parameter at a time and keep all other settings fixed.
As shown in Fig.~\ref{fig:hyper_appendix}, most curves follow a U-shaped trend and reach their minimum near 10 on both datasets.
Small values provide insufficient coverage of plausible destinations or receiver-side community evidence, which is particularly visible for Radius and Flow-EMD.
Conversely, increasing either parameter to 15 or 20 introduces weakly relevant candidates and locations, degrading Comm-Size and Motif consistency and increasing reasoning cost.
The similar behavior across individual, aggregate, and structural metrics indicates that focused evidence is more useful than an unrestricted reasoning context.
We therefore use \texttt{top\_k}=10 and \texttt{active\_m}=10 in all main experiments.

\section{Agent Prompt Templates}
\label{appendix:prompts}
\begin{figure*}[t]
    \centering
    \includegraphics[width=0.8\textwidth]{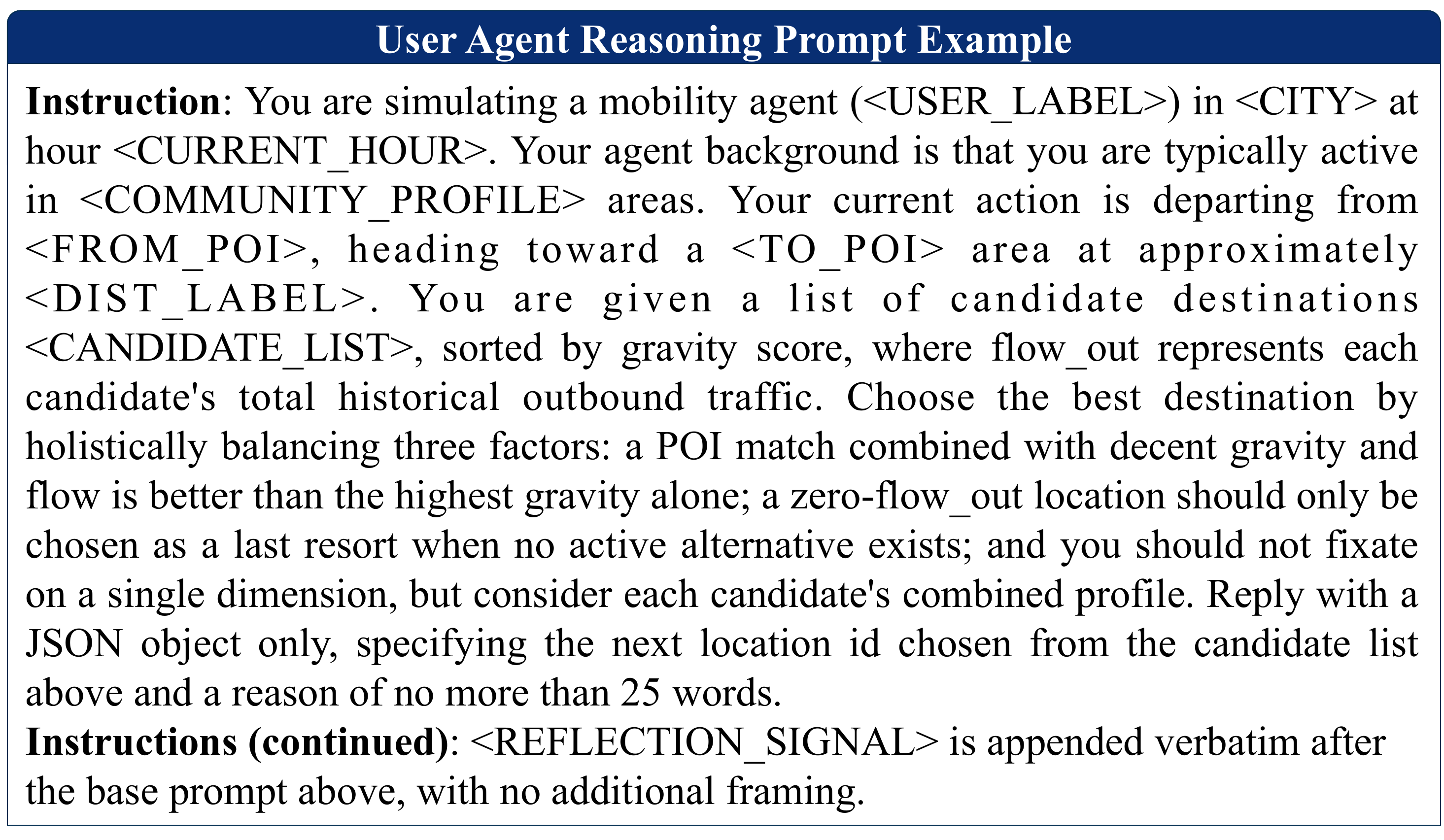}
    \caption{Prompt template for individual-view destination reasoning by a user agent.}
    \label{fig:user_prompt}
\end{figure*}

\begin{figure*}[t]
    \centering
    \includegraphics[width=0.8\textwidth]{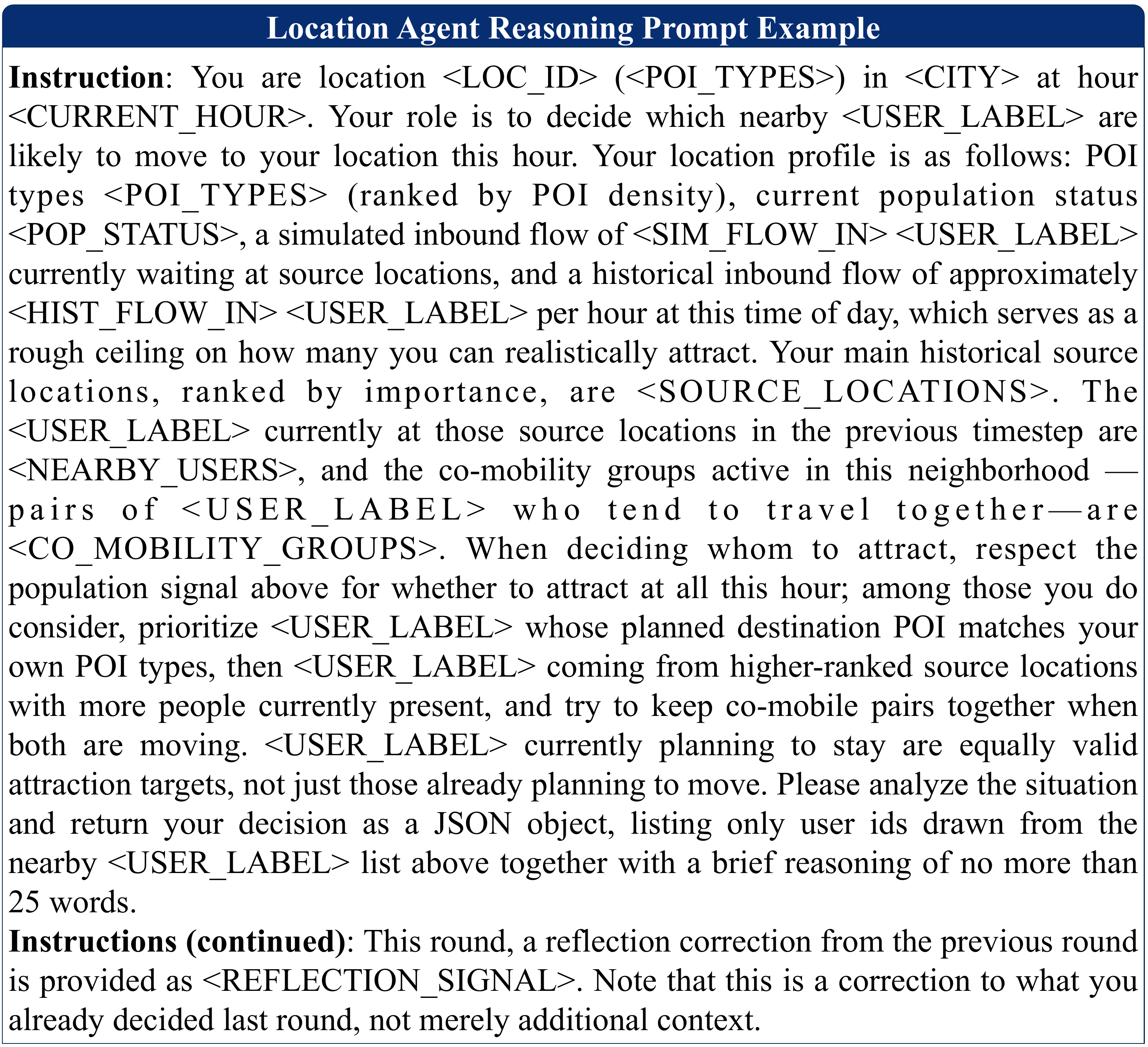}
    \caption{Prompt template for receiver-side community reasoning by a location agent.}
    \label{fig:location_prompt}
\end{figure*}

\begin{figure*}[t]
    \centering
    \includegraphics[width=0.8\textwidth]{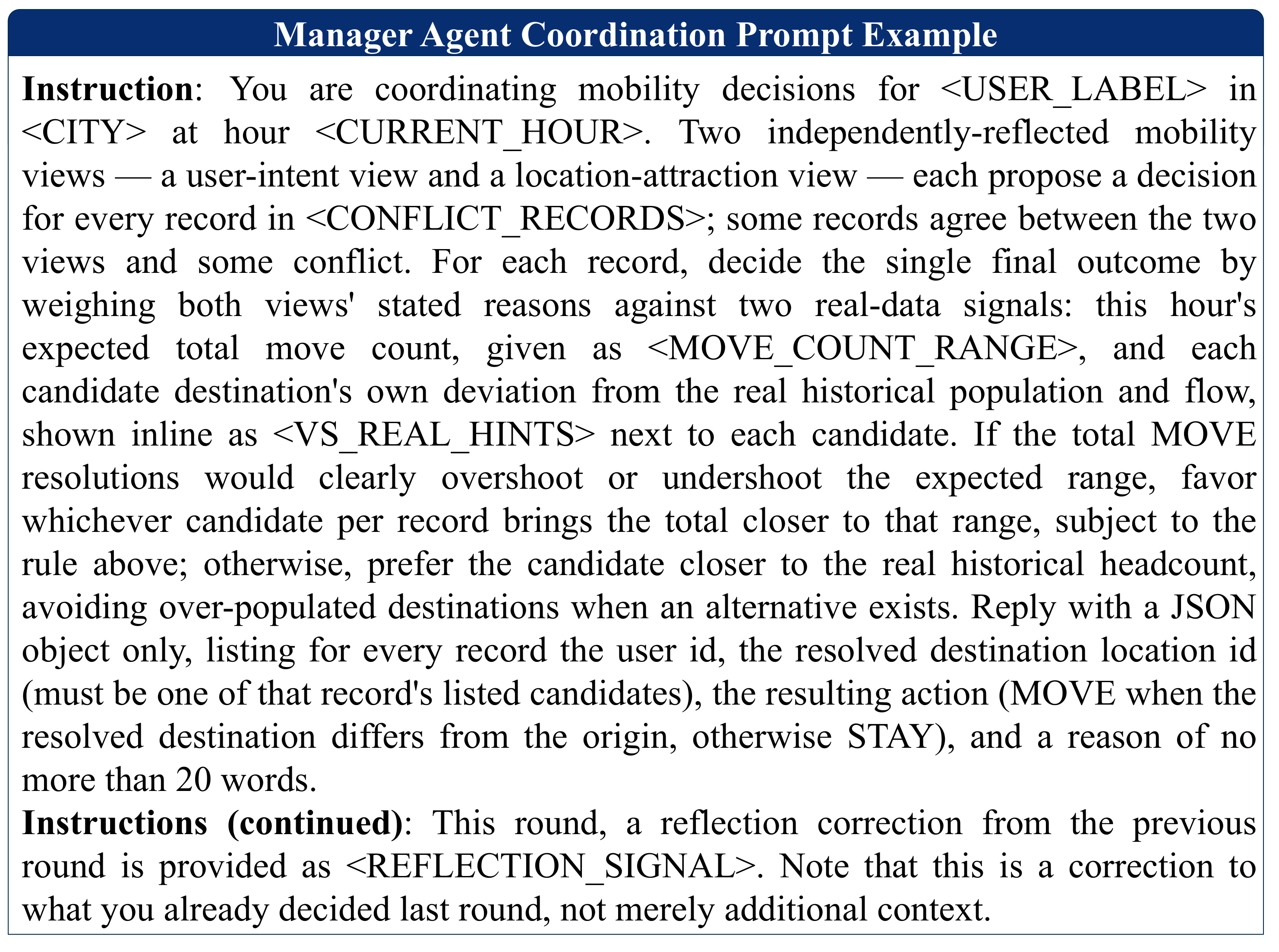}
    \caption{Prompt template for individual-community coordination by the manager agent.}
    \label{fig:coordination_prompt}
\end{figure*}

\begin{figure*}[t]
    \centering
    \includegraphics[width=0.8\textwidth]{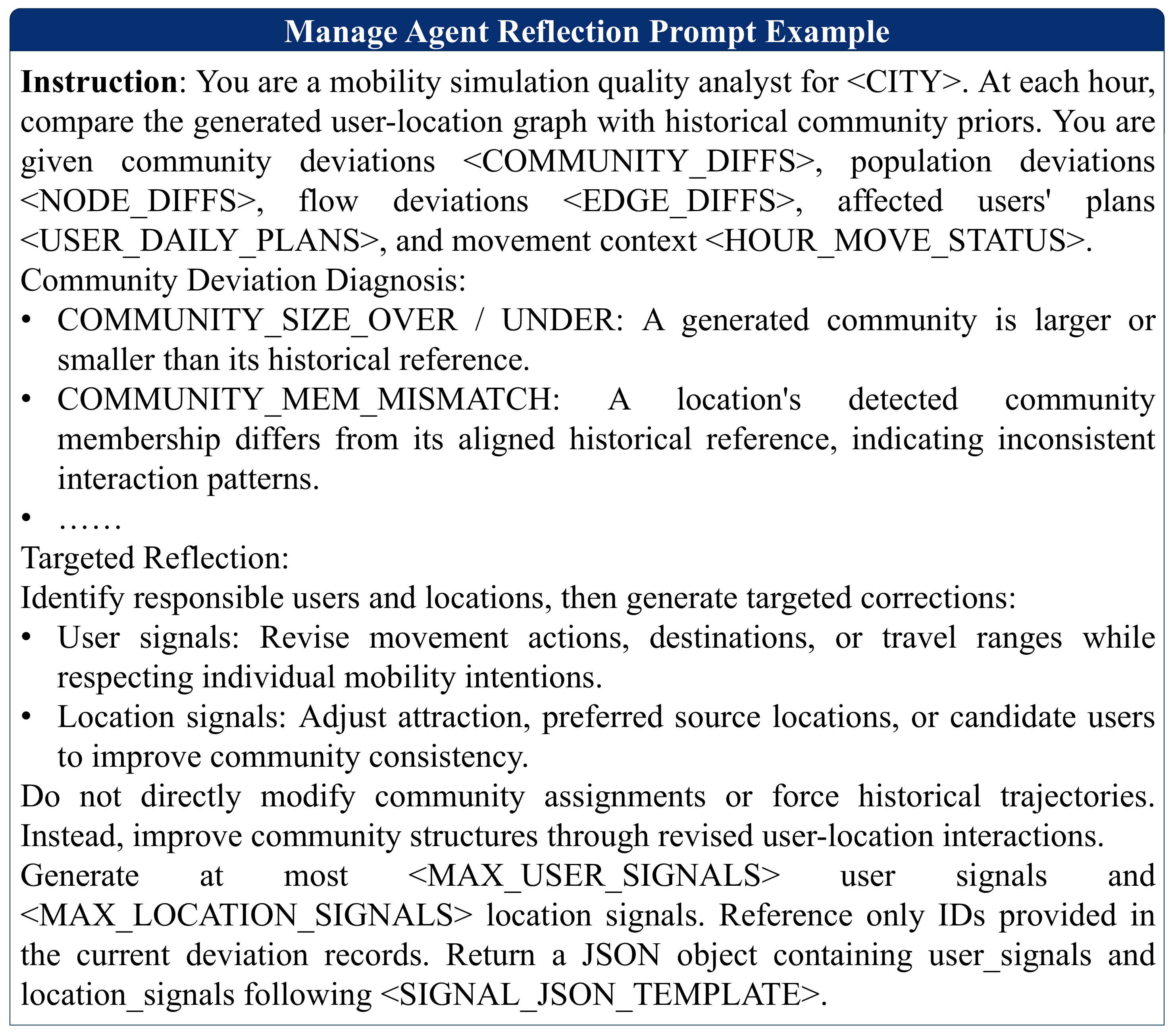}
    \caption{Prompt template for community-consistency reflection by the manager agent.}
    \label{fig:reflection_prompt}
\end{figure*}

Figs.~\ref{fig:user_prompt}--\ref{fig:reflection_prompt} present the complete prompt templates used by the three agent roles.
All templates follow a shared structure consisting of a role definition, time-step context, structured mobility evidence, decision constraints, and a JSON-only output schema.
Bracketed fields are instantiated at runtime using the current mobility state and community-conditioned interaction priors, ensuring that agents receive only evidence relevant to the current graph update.

The user-agent prompt specifies the user's current location, planned POI semantics and distance range, community profile, and retrieved destination candidates with attraction and flow evidence.
The location-agent prompt provides the target location's semantic function, population status, historical and simulated incoming flows, connected source locations, nearby users, and active co-mobility groups.
These two prompts generate complementary destination-side and receiver-side proposals rather than asking a single agent to reason over the entire evolving graph.

The manager-agent coordination prompt presents only records for which the two views require resolution, together with expected movement volume and deviations from historical population-flow patterns.
The reflection prompt summarizes node-population and edge-flow discrepancies, affected users' daily plans, and temporal movement conditions, and requests bounded correction signals for the implicated user and location agents.
Reflection signals are appended to subsequent agent prompts as explicit corrections rather than unconstrained additional context.
Restricting every response to a predefined JSON schema enables deterministic parsing and identifier validation against the supplied candidate sets, and prevents free-form outputs from directly modifying the graph state.

\end{document}